\documentclass[10pt,twocolumn,letterpaper]{article}

\usepackage[pagenumbers]{iccv} % To force page numbers, e.g. for an arXiv version

\usepackage{algorithm}
\usepackage{algpseudocode}
\usepackage{setspace}
\usepackage{multicol}
\usepackage{multirow}
\usepackage{tabularx}
\usepackage{booktabs}
\usepackage{makecell}
\usepackage{colortbl}
\usepackage[table]{xcolor}
\usepackage{bm}

\definecolor{gray}{RGB}{220,220,220}
\definecolor{lgreen}{RGB}{50,205,50}
\definecolor{ored}{RGB}{255,69,0}

\usepackage{soul}
\definecolor{iccvblue}{rgb}{0.21,0.49,0.74}
\usepackage[pagebackref,breaklinks,colorlinks,allcolors=iccvblue]{hyperref}

\def\paperID{10255} % *** Enter the Paper ID here
\def\confName{ICCV}
\def\confYear{2025}

\title{CCDNet: Learning to Detect Camouflage against Distractors \\in Infrared Small Target Detection}

\author{Zikai Liao\\
Stony Brook University\\
\and
Zhaozheng Yin\\
Stony Brook University\\
}

\begin{document}
\maketitle
\begin{abstract}
% The ABSTRACT is to be in fully justified italicized text, at the top of the left-hand column, below the author and affiliation information.
% Use the word ``Abstract'' as the title, in 12-point Times, boldface type, centered relative to the column, initially capitalized.
% The abstract is to be in 10-point, single-spaced type.
% Leave two blank lines after the Abstract, then begin the main text.
% Look at previous \confName abstracts to get a feel for style and length.

Infrared target detection (IRSTD) tasks have critical applications in areas like wilderness rescue and maritime search. However, detecting infrared targets is challenging due to their low contrast and tendency to blend into complex backgrounds, effectively camouflaging themselves. Additionally, other objects with similar features (distractors) can cause false alarms, further degrading detection performance. To address these issues, we propose a novel \textbf{C}amouflage-aware \textbf{C}ounter-\textbf{D}istraction \textbf{Net}work (CCDNet) in this paper. We design a backbone with Weighted Multi-branch Perceptrons (WMPs), which aggregates self-conditioned multi-level features to accurately represent the target and background. Based on these rich features, we then propose a novel Aggregation-and-Refinement Fusion Neck (ARFN) to refine structures/semantics from shallow/deep features maps, and bidirectionally reconstruct the relations between the targets and the backgrounds, highlighting the targets while suppressing the complex backgrounds to improve detection accuracy. Furthermore, we present a new Contrastive-aided Distractor Discriminator (CaDD), enforcing adaptive similarity computation both locally and globally between the real targets and the backgrounds to more precisely discriminate distractors, so as to reduce the false alarm rate. Extensive experiments on infrared image datasets confirm that CCDNet outperforms other state-of-the-art methods. 
% The code will be released.

\end{abstract}    
\vspace{-1.5em}
\section{Introduction}

\begin{figure}
    \centering
    \includegraphics[width=\linewidth,keepaspectratio]{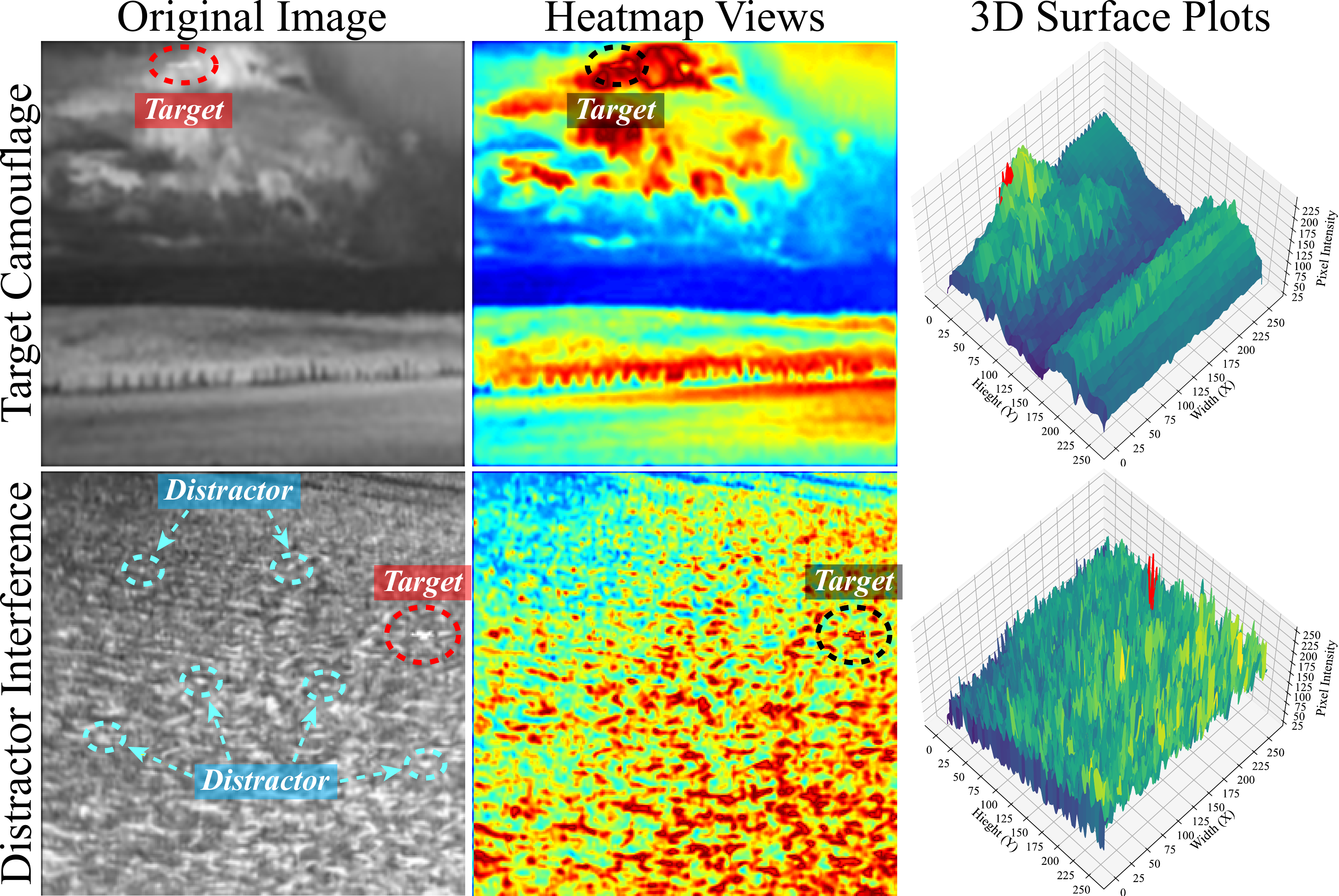}
    \vspace{-1.5em}
    \caption{Illustrations of two key challenges in IRSTD tasks: target camouflage and distractor interference.}
    \vspace{-1.5em}
    \label{fig:challenges}
\end{figure}

Infrared imaging technology has been widely applied to detection tasks for the past decades, such as traffic management, wilderness rescue, and marine search \cite{wang2019mdvsfa,dai2021acm,dai2021alcnet,zhang2022isnet,luo2022imnn,kou2023infrared_survey,fang2023dagnet,fang2023danet,liu2024sls,fang2024scinet}. Numerous methods have been proposed for these wide applications, broadly categorized into traditional and deep-learning (DL) approaches. Traditional methods \cite{zhang2019infrared, sun2020infrared,liu2021nonconvex,luo2022imnn, liu2023infrared, wang2024infrared,huang2024infrared} primarily rely on handcrafted feature filters that exploit background-foreground differences to detect targets, while the DL methods \cite{wang2019mdvsfa,dai2021acm,dai2021alcnet,li2022dnanet,zhang2022isnet,fang2023dagnet,fang2023danet,liu2024sls,zhang2024irprunedet,fang2024scinet} employ systematic semantic learning to adaptively model rich target feature representations, demonstrating superior performance in IRSTD tasks. However, infrared images typically exhibit lower resolution, poorer quality, and heavier noise and clutter compared to visible images \cite{dai2021acm,zhang2022isnet,fang2023dagnet,fang2023danet,fang2024scinet}, posing two major unresolved challenges for infrared small target detection (IRSTD): \textbf{(1) Target camouflage}. Targets in infrared images, especially small-scale ones, are usually low in contrast, blurry in contour, and deficient in texture \cite{dai2021acm,fang2023dagnet,fang2023danet,fang2024scinet}, which often blend into complex backgrounds, become camouflaged and hard to identify; \textbf{(2) Distractor interference}. Complex backgrounds may contain objects with features similar to the real targets, which might lead to a high false alarm rate \cite{wang2019mdvsfa,fang2023danet}. \cref{fig:challenges} displays two infrared images and their visualizations to demonstrate these challenges. Many of the abovementioned existing approaches still struggle to effectively addressing these two key challenges, leading to suboptimal detection performance in real-world applications.

In this paper, we propose to solve the above-mentioned two key challenges from the aspects of camouflage object detection and contrastive learning, leading to a novel IRSTD method to detect camouflaged targets while being robust to distractor inferences, namely \textbf{C}amouflage-aware \textbf{C}ounter-\textbf{D}istraction \textbf{Net}work (CCDNet).
% in this paper, we design a novel IRSTD method to detect camouflaged targets robust to distractor interferences, namely Camouflage-aware Counter-Distraction Network (CCDNet), leveraging ideas from camouflage object detection and contrastive learning to effectively improve the performance of the IRSTD task. 

Existing works on camouflage object detection primarily focus on relatively large-scale animals in visible images \cite{fan2020sinet,li2021ujscod,he2023feder,jia2022segmar}, leveraging deep networks
% with stacked convolutions to enhance receptive field information 
for contour discovery. However, infrared targets, especially small-scale ones, exhibit comparatively sparse features with limited semantic richness, making them particularly vulnerable to overly deep networks due to spatial downsampling
% and are likely to \textit{suffer from a network too deep} due to spatial downsampling 
\cite{dai2021acm,dai2021alcnet,fang2023dagnet}.
% where they utilize relatively deep networks with stacks of convolutions to obtain rich receptive field information for object contour discovery. 
% However, a network too deep may result in losing vital infrared target features due to downsampling operations \cite{dai2021acm,dai2021alcnet,fang2023dagnet}, and infrared targets usually contain very sparse features (especially small-scale targets).
% won't yield sufficient information favorable for detection.
% and won't yield favorable target information since infrared targets are relatively small with very sparse features \cite{fang2023dagnet,fang2023danet}.
% which is harmful to IRSTD tasks. 
% Moreover, unlike animals that usually occupy a great portion of natural images, infrared targets are relatively small, which means stacking up convolutions won't yield very useful target information. 
To this end, we propose to extend the backbone's width instead of its depth, and enable the joint exploitation of target and background features.
% to enhance feature representation learning.
% and exploit the rich semantics of the targets and backgrounds to \textit{jointly guide feature representations} of the targets.
% and take advantage of semantics/structures from deep/shallow layers to cross-guide the feature representations of targets and their backgrounds.
We first assemble a series of proposed Weighted Multi-branch Perceptrons (WMPs) in the backbone to extract rich contextual features, regulated by adaptive self-conditioning. In addition, we design an Aggregation-and-Refinement Fusion Neck (ARFN) with Top-down Background Semantic Guidance (TBSG) and Bottom-up Object Structure Enhancement (BOSE). The TBSG takes deep features as background semantics, and inversely merges with shallow features to remove irrelevant structures in the background. In contrast, the BOSE excavates crucial structures from shallow features as contours, and merges them with deep features to refine semantic distributions. Our ARFN
% joint use of TBSG and BOSE in ARFN
exchanges vital visual cues so target features can be implicitly highlighted. With WMPs and ARFN, our CCDNet can more effectively identify camouflages
% camouflaged targets
in complex backgrounds.

% Firstly, unlike previous COD works that extend the backbone network's depth to aggregate rich and deep receptive field information for boundary discovery of camouflaged objects \cite{fan2020sinet}, we argue that a network too deep can be harmful to preserving vital features of infrared targets, especially small ones \cite{dai2021acm,fang2023dagnet}. Instead, we assemble a series of proposed Weighted Multi-branch Perceptrons (WMP)

Contrastive Learning has been widely employed in vision tasks \cite{chen2020simclr,he2020moco, xie2021detco,wang2022contrastmask,yu2022towards,qi2024imc}, but not to discriminate distractors \cite{fang2023danet}.
% not popularly explored to discriminate real targets and their distractors \cite{fang2023danet}.
Therefore, we propose a novel Contrastive-aided Distractor Discriminator (CaDD) to achieve this goal, which includes a Local Contrastive Module (LCM) and a Global Contrastive Module (GCM). The LCM reevaluates the relations between the target and its neighboring regions via a triple difference assessment, and imposes saliency enhancement on the target region.
% against its neighbors.
% regional weighting with three branches of feature differences, 
% and establishes a loss to maximize the differences between the old regions and the new ones. 
% The LCM will not only contribute to network training, but also serve as a feature enhancement module that further facilitates accurate target detection. 
The GCM
% on the other hand, 
automatically selects distractor regions as negative samples across iterations, and enforces similarity computation between targets and distractors, explicitly penalizing the network training to facilitate the discrimination of them.
% adding an explicit penalty to network training that facilitates the discrimination of them.
% and adopts a contrastive loss that improves discrimination on distractors of the proposed DICNet through network training. 
With LCM and GCM, our CCDNet can be more robust to distractor interference.

We conduct experiments on three infrared datasets, and the results verify the superiority of our proposed method compared to other state-of-the-art (SOTA) methods. Our contributions are summarized as follows:

$\bullet$ To effectively detect camouflaged targets, we design a Weighted Multi-branch Perceptron (WMP) backbone to capture rich contextual features.
% that facilitate accurate recognition of targets and backgrounds, 
Additionally, we develop the Aggregation-and-Refinement Fusion Network (ARFN), which integrates two feature refinement modules, leveraging rich semantics and structures for effective camouflaged target identification.

$\bullet$ To enhance the discrimination of distractors, we present a novel Contrastive-aided Distractor Discriminator (CaDD), which reduces false alarms by actively learning feature distinctions between targets and distractors through local and global contrastive learning.
% which learns feature differences between targets and distractors via local and global active contrastive learning, and thus reduces false alarms.
% to jointly discriminate distractors in the images through active contrastive learning.

% $\bullet$ We propose a novel approach to address IRSTD challenges, i.e., incorporating ideas from COD and CL into IRSTD tasks.
$\bullet$ 
% To tackle the challenges of detecting camouflaged infrared targets
% % in complex backgrounds 
% and discriminating distractors, 
With WMP, ARFN and CaDD, we come up with the Camouflage-aware Counter-Distraction Network (CCDNet) for IRSTD tasks. Extensive experiments on IRSTD datasets verify that our CCDNet can detect infrared targets with high performance and real-time speed, outperforming other state-of-the-art (SOTA) methods.
% and employ contrastive learning that suppresses distraction interferences. 
% Extensive experiments have shown superior results compared to other state-of-the-art (SOTA) detection methods.

\begin{figure*}[t]
    \centering
    \includegraphics[width=\linewidth,keepaspectratio]{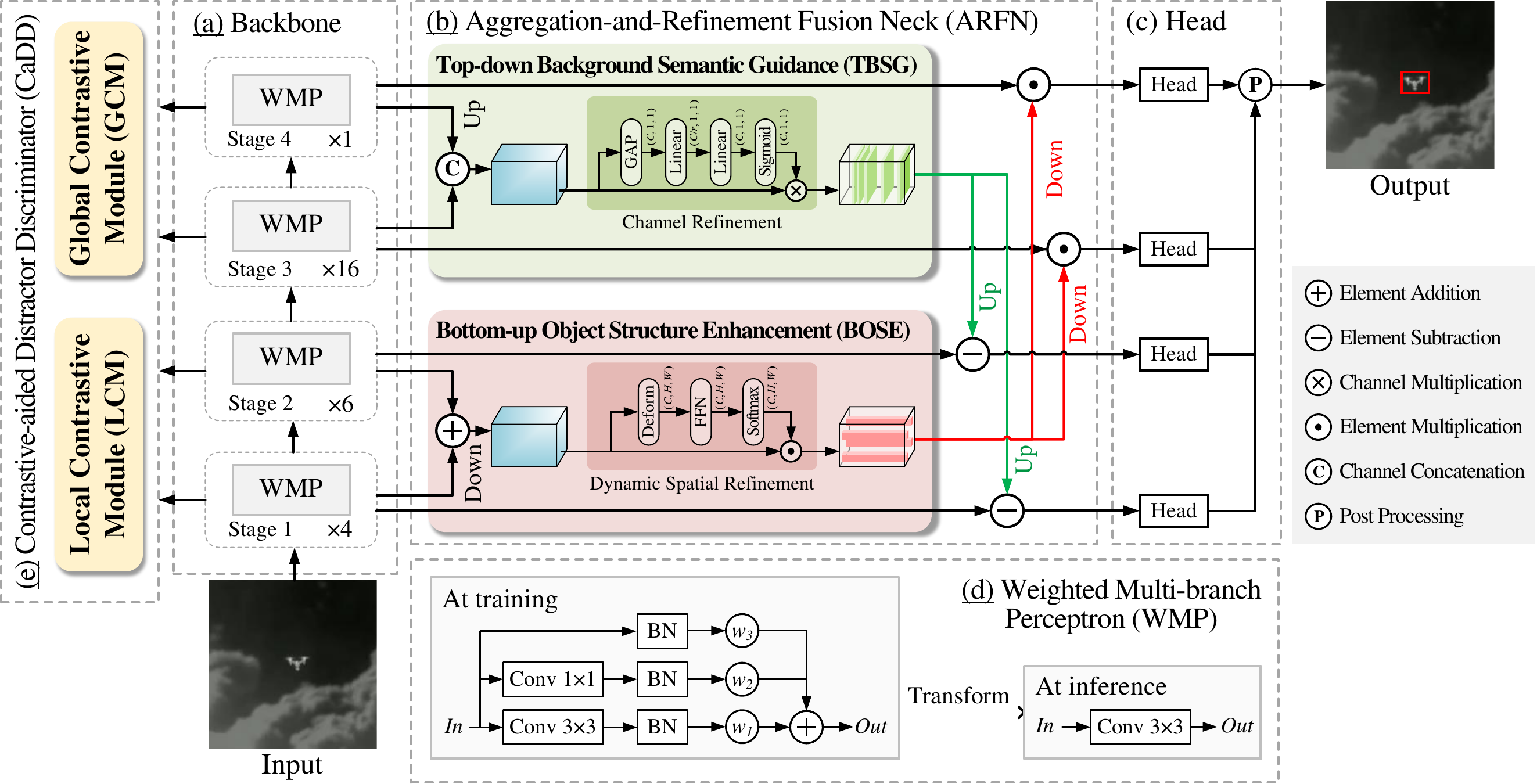}
    \vspace{-1.6em}
    \caption{Overview of our proposed CCDNet. It has a backbone with proposed WMPs to extract rich contextual features, and ARFN with proposed TBSG and BOSE to highlight target features for camouflage detection via adaptive guidance from key semantics and structures. The proposed CaDD, with LCM and GCM, enables the network to better differentiate between real targets and their distractors.} 
    % The WMP has three branches during training and is transformed into a single branch during inference.}
    \vspace{-1.5em}
    \label{fig:ccdnet-overview}
\end{figure*}

\vspace{-0.5em}
\section{Related Works}
\vspace{-0.2em}

\subsection{Infrared Small Target Detection}
\vspace{-0.2em}

Traditional IRSTD methods, like low-rank methods \cite{zhang2019infrared,liu2021nonconvex,kong2021infrared,pang2024lrtasp} and filter-based methods \cite{zhu2020balanced,zhang2023infrared,gao2024infrared}, mostly rely on handcrafted feature priors to identify the targets, thus not robust to complex backgrounds and strong interferences \cite{dai2021acm,dai2021alcnet,zhang2022isnet,fang2023dagnet}.
% \textbf{XXXXX} \cite{zhang2019infrared, sun2020infrared,liu2021nonconvex,luo2022imnn, liu2023infrared, wang2024infrared,huang2024infrared}. 
% However, they usually are not robust in dealing with complex backgrounds and strong interferences. 
Meanwhile, DL methods can adaptively learn rich target feature representations with versatile architectures, often delivering superior performance. Recent works \cite{wang2019mdvsfa,dai2021acm,li2022dnanet,zhang2022isnet,fang2023dagnet,fang2023danet,ying2023iesps,zhang2024irprunedet,liu2024sls} have presented great detection results on IRSTD tasks, but they pay less attention to targets that are seemingly camouflaged in their backgrounds, and are comparatively liable to distractor interference. In this paper, we propose the CCDNet to address these issues by restructuring the target and background features for better detection performance.
% we propose to incorporate ideas from COD and CL to address this issue.
% , instead, explores a new idea of regarding the infrared target detection task as a special camouflaged object detection task, and incorporating it into our network architecture, so it can effectively detect targets blended into complex backgrounds to realize early discovery or take further actions. \textbf{NEED MORE}
% To the best of our knowledge, this is the first time

\vspace{-0.2em}
\subsection{Camouflage Object Detection}
\vspace{-0.2em}

Camouflage object detection aims to identify objects with features almost matched to their backgrounds via strong visual perception designs \cite{fan2020sinet}. Current methods, such as \cite{fan2020sinet,li2021ujscod,jia2022segmar,zhai2022fganet,he2023feder,zhao2024lakered}, emphasize exploiting rich semantics via aggregating complicated receptive field information, which usually requires to deepen the network.
% Recent DL-based methods have been utilizing some intrinsic strategies to solve this issue. 
% SINet \cite{fan2020sinet}, proposed by Fan \etal, an improved U-Net architecture with rich receptive field information aggregations and extractions. Li \etal proposed UJSCOD \cite{li2021ujscod}, which jointly detects salient and camouflaged objects with a novel feature encoder-decoder structure to finely extract and differentiate features of different tasks. Jia \etal designed SegMaR \cite{jia2022segmar}, achieving detection while magnifying camouflaged feature representations in parallel across network stages. Zhai \etal presented FGA-Net \cite{zhai2022fganet}, within which introduces a novel Figure-Ground Aided module utilizing crucial visual cues to separate foregrounds and backgrounds. He \etal proposed FEDER \cite{he2023feder}, featuring wavelet decomposed features and boundary information to discover hidden objects. Zhao \etal designed LAKE-RED \cite{zhao2024lakered} incorporating diffusion model to facilitate a finer identification of camouflaged objects. Pang \etal proposed ZoomNeXt \cite{pang2024zoomnext}, integrating mixed-scale features by zooming in/out object regions to excavate important visual clues. 
However, most of them only apply to natural visible images, where targets are often larger and contain more feature details unlike in IRSTD tasks.
% that don't apply to IRSTD tasks.
% and their deep network architectures might not be effective for IRSTD tasks, particularly when targets are small, and may be severely affected by complex backgrounds and interfering clutter.
Our work adaptively expands the network's width instead of depth to avoid feature collapsing by over-downsampling, and leverages rich semantics of backgrounds and rich structures of targets to bidirectionally fuse semantics and structures that form a target-centric representation favorable for accurate detection.

\vspace{-0.5em}
\subsection{Contrastive Learning}
\vspace{-0.2em}

Contrastive learning (CL) methods have been proven effective 
% for various vision tasks 
by enforcing similarity computation on positive/negative samples to implicitly instruct models to discriminate foreground/background features \cite{chen2020simclr,he2020moco, xie2021detco,wang2022contrastmask,yu2022towards,qi2024imc}. Nevertheless, employing CL to discriminate infrared targets and distractors is not widely explored.
% , due to issues of negative sample definition. 
\cite{fang2023danet} randomly selects background regions as distractors as negative samples to perform CL, which contradicts the definition of distractors. Another intuitive way to realize this is to manually label distractors as negative samples, but this is laborsome and may introduce biases. In our work, we design the CaDD with two contrastive learning modules: The LCM treats the target's neighboring regions as negative and enforces contrastive loss to enhance the target saliency, and the GCM improves distractor discrimination by computing similarity between real targets and distractors.
% those detection results with lower confidence scores. 
The joint use of them can reduce the false alarm rate, thus improving detection accuracy.
\vspace{-0.8em}
\section{Method}
\vspace{-0.4em}

% In this section, we first introduce the overview of our CIDNet (Sec.\ref{sec:3.1}). Then, we present the details of our proposed Weighted Multi-branch Perceptron (WMP) (Sec.\ref{sec:3.2}) and the Aggregation-and-Refinement Fusion Neck (ARFN) with Top-down Background Semantic Guidance (TBSG) and Bottom-up Object Structure Enhancement (BOSE) (Sec.\ref{sec:3.3}), followed by the design of our Contrastive-aided Distractor Discrimination (CaDD) (Sec.\ref{sec:3.4}) and the loss function (Sec.\ref{sec:3.5}). More details on the implementation of our proposed method are in the supplementary material.

\subsection{Overview of CCDNet} \label{sec:3.1}

\cref{fig:ccdnet-overview} displays the overview of our CCDNet. It uses a backbone (\cref{fig:ccdnet-overview}(a)) composed of four stages of the proposed WMPs (\cref{fig:ccdnet-overview}(d)) to extract and aggregate rich contextual features for accurate target feature representations. Then, it forwards multi-stage features to our ARFN (\cref{fig:ccdnet-overview}(b)), which exploits fine structures and coarse semantics from shallow and deep feature maps, respectively, and jointly refines target features for better camouflage identification. At last, refined features are fed into the detection heads (\cref{fig:ccdnet-overview}(c)) to output classification and localization results. During training stage, backbone output features will be used in our proposed CaDD (\cref{fig:ccdnet-overview}(e)), which enforces local and global similarity computation to guide our CCDNet to differentiate real targets and distractors.
% the backbone, broken down into four stages with various numbers of proposed WMPs, extracts and aggregates rich multi-level contextual features that empower accurate target detection; the ARFN, leveraging the proposed TBSG and BOSE modules to exploit key structures and semantics from shallow and deep feature maps, and guide different feature representations to highlight the targets; the head, responsible for classification and localization. Besides, the proposed CaDD takes output feature maps from different backbone stages, and enforces similarity computation both locally and globally, directing the network to differentiate the real targets and their distractors. 
% \ul{Implementation details of our proposed method can be found in the supplementary material in }\cref{sec:x-implement}.

\vspace{-0.5em}
\subsection{Weighted Multi-branch Perceptron Backbone} \label{sec:3.2}

Many camouflage object detectors tend to enlarge the receptive field by deepening their architecture to aggregate rich feature information for boundary discovery. To avoid downsampling operations in deep architectures collapsing spatial feature integrity of the infrared targets (especially for small-scale ones) \cite{dai2021acm,dai2021alcnet,fang2023dagnet,fang2023danet}, 
% inspired by RepVGG \cite{ding2021repvgg}, 
we present the WMP block in our backbone (\cref{fig:ccdnet-overview}(d)), which expands the network's width instead of depth, with three branches capturing contextual features of different receptive fields respectively, and finally merge information from these branches to form fused features. However, we argue that each branch may contribute differently to represent target features. Therefore, we propose an adaptive self-conditioning mechanism that directs the feature fusion in WMP, where each branch is controlled by a learnable conditioning parameter to adaptively decide the branch importance. By this, the WMP fused output can represent target features more accurately, thus facilitating detection performance. Furthermore, we adopt a network pruning strategy similar to \cite{ding2021repvgg} to reduce model complexity while maintaining performance, details of which can be found in \ref{sec:x-wmp} of the supplement.
% % which expands the network's width by extracting multi-level contextual features with rich receptive field information via three different branches, 
% controlled by an adaptive self-conditioning mechanism that directs the merge of multi-level feature extraction. As shown in \cref{fig:ccdnet-overview} (b) and (d), the input feature map will go through three branches,
% % a Conv\_3x3 branch, a Conv\_1x1 branch, and a residual branch, 
% and each branch is controlled by an adaptive self-conditioning learnable parameter to decide its importance. After all three branches, the outputs will be element-wisely added as the final output.

% By using WMPs, the backbone can extract rich receptive field information that can effectively facilitate identifying camouflage targets within complex backgrounds, without introducing too many downsampling operations. 
% By using WMPs, 
% It's worth noticing that our WMP has three branches during training, but will transform into a single branch during inference, which not only saves computation but also maintains the model's detection accuracy \cite{ding2021repvgg}. 
% \ul{The transformation process can be found in the supplement} \cref{sec:x-wmp}.

\subsection{Aggregation-and-Refinement Fusion Neck} \label{sec:3.3}

Due to downsampling operations and small target sizes, the backbone usually extracts rich target structures at shallow stages but very few semantics at deep stages \cite{dai2021acm,dai2021alcnet,fang2023dagnet,fang2023danet}, where semantics from deep layers are usually dominated by the complex backgrounds. In our work, we design a novel ARFN (shown in \cref{fig:ccdnet-overview}(c)) with two different feature fusion modules: TBSG to fully utilize the background semantics from deep stages to inversely guide the target structures in the shallow stages, and BOSE which leverages fine structures from shallow stages to refine the high-level semantics in the deep stages. The joint use of TBSG and BOSE
can form a target-centric feature representation, thus improving detection performance on camouflaged targets.

\vspace{-0.2em}
\subsubsection{Top-down Background Semantic Guidance}

Our proposed TBSG (\cref{fig:ccdnet-overview}(b)) focuses on feature maps from deep stages with rich background semantics, and use them to inversely enrich target semantics at shallow stages and further guide the 
% precise 
representations of target structures. 

We first take the two output feature maps from stages 3 ($\mathbf{F}_{3}$) and 4 ($\mathbf{F}_{4}$) of the backbone, upsample $\mathbf{F_{4}}$ 2x and concatenate with $\mathbf{F}_{3}$, and use a channel refinement \cite{hu2020senet} to form the enhanced background-related channel semantics $\mathbf{F}^{\text{bkg}}$. After this, we upsample $\mathbf{F}^{\text{bkg}}$ 2x and 4x, respectively, and negatively sum up with feature maps from stages 2 
 ($\mathbf{F}_{2}$) and 1 ($\mathbf{F}_{1}$), respectively. This process is expressed as:
\vspace{-0.2em}
\begin{align}
    \mathbf{F}^{\text{bkg}}&=\text{CR}(\text{Cat}[\mathbf{F}_3;\text{Up}_{\text{2x}}(\mathbf{F}_4)]), \\
    \mathbf{F}_2^{\text{out}}&=\mathbf{F}_2+(-\text{Up}_{\text{2x}}(\mathbf{F}^{\text{bkg}})), \\
    \mathbf{F}_1^{\text{out}}&=\mathbf{F}_1+(-\text{Up}_{\text{4x}}(\mathbf{F}^{\text{bkg}})),
\end{align}

\vspace{-0.4em}
\noindent where Up($\cdot$) means upsampling, 2x and 4x denote upsampling ratios, Cat[;] is concatenation, and CR($\cdot$) is channel refinement.
% $F_i$ is the $i$-th stage feature map from the backbone, 
$\mathbf{F}_1^{\text{out}}$ and $\mathbf{F}_2^{\text{out}}$ are the final output feature map from the neck at the 1st and 2nd stage. In this way, the inverse semantics can serve as the visual cue that not only guides what the background contents are, but also suppresses irrelevant background features by subtracting with shallow feature maps, thus improving the network's awareness of the targets and further benefiting accurate target detection.

\vspace{-0.2em}
\subsubsection{Bottom-up Object Structure Enhancement}

With crucial target features in the shallow feature maps, our BOSE (\cref{fig:ccdnet-overview}(c)) exploits
% takes advantage of 
those target structures to refine the semantic representations in the deep feature maps. 

Similar to TBSG, feature maps from the first two stages $\mathbf{F}_{1}$ and $\mathbf{F}_{2}$ are used to compose comprehensive target structure features, with $\mathbf{F}_{2}$ element-wisely adding the 2x downsampled $\mathbf{F}_{1}$. We then feed this summed feature map into the dynamic spatial refinement to precisely enhance the target regions without feature compression. 
% The structure of the dynamic spatial refinement is shown in \cref{fig:ccdnet-overview}(c). 
The 3x3 deformable convolution \cite{zhu2019dcnv2} is used to excavate non-local multi-scale spatial features of the targets, and the feed-forward network (FFN) for channel projection. The final spatial-refined weight $\mathbf{F}^{\text{tgt}}$ is generated by softmax activation and imposed on the summed feature map. 
% After spatial enhancement, 
Afterwards, $\mathbf{F}^{\text{tgt}}$ is downsampled 2x and 4x to be element-wisely multiplied on $\mathbf{F}_{3}$ and $\mathbf{F}_{4}$. The process of BOSE is formulated as:
\vspace{-0.5em}
\begin{align}
    \mathbf{F}^{\text{tgt}}&=\text{DSR}(\mathbf{F}_2+\text{Down}_{\text{2x}}(\mathbf{F}_1)), \\
    \mathbf{F}_3^{\text{out}}&=\mathbf{F}_3\odot\text{Down}_{\text{2x}}(\mathbf{F}^{\text{tgt}}), \\
    \mathbf{F}_4^{\text{out}}&=\mathbf{F}_4\odot\text{Down}_{\text{4x}}(\mathbf{F}^{\text{tgt}}),
\end{align}

\vspace{-0.5em}
\noindent where Down($\cdot$) means downsampling, 
% 2x and 4x denote downsampling ratios, 
$\odot$ is element-wise multiplication, and DSR($\cdot$) is dynamic spatial refinement. With the proposed BOSE, the structure information from shallow layers can guide the representations of sementics in deep layers, and therefore highlight target features in deep layers to avoid them being overwhelmed by background features.
% the semantics from deep layers can perceive how structures are distributed across the image, and therefore reconstructing high-level features to better represent the target features against complex backgrounds.

\subsection{Contrastive-aided Distractor Discrimination} \label{sec:3.4}

\begin{figure*}[t]
    \centering
    \includegraphics[width=\linewidth,keepaspectratio]{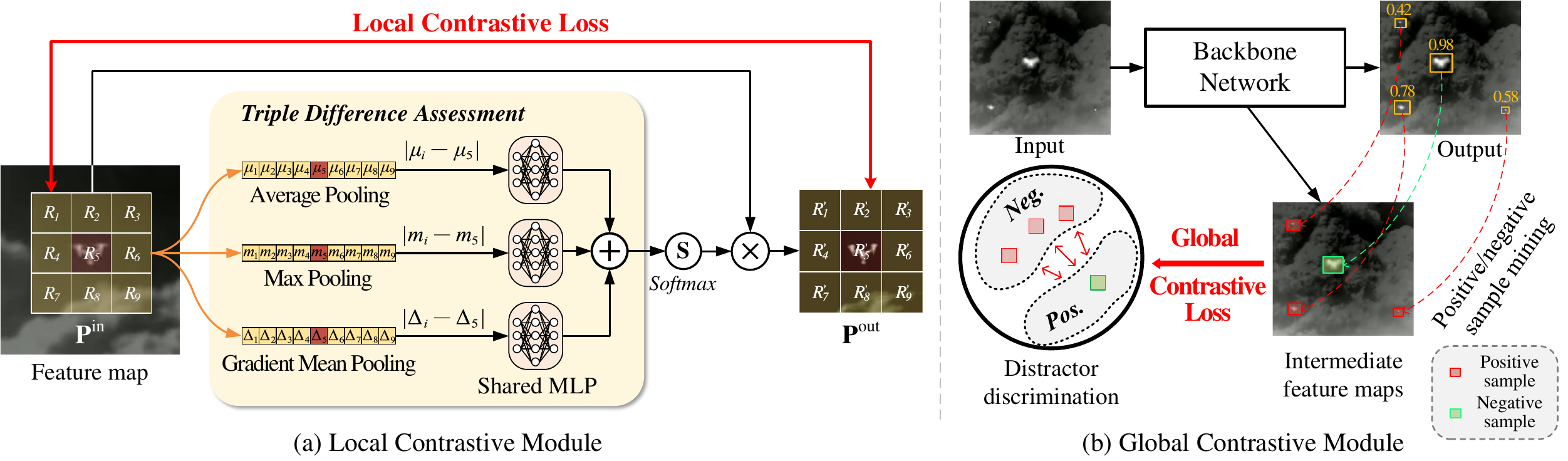}
    \vspace{-2em}
    \caption{Illustration of our proposed LCM and GCM in CaDD. $\mathbf{P}^{\text{in}}$ and $\mathbf{P}^{\text{out}}$ denote the nine-region area from the input feature map and the LCM-processed feature map. \textit{Pos.} and \textit{Neg.} means the positive and negative sample. Both LCM and GCM apply to all four output feature maps from the backbone. For visualization purposes, we only display its mechanism with only one feature map.}
    \vspace{-1.5em}
    \label{fig:cadd}
\end{figure*}

In real-world applications, some objects in the complex background might resemble the real targets, and might lead to false alarms. To address this issue, we devise CaDD with two main modules, LCM and GCM (\cref{fig:cadd}), to adaptively improve the network's ability to discriminate distractors in complex backgrounds with active contrastive learning. Note that our CaDD is only employed during training, thus having no computation overhead for inference.

% Note that CaDD is used on all four output feature maps from the backbone for contrastive learning (as shown in \cref{fig:ccdnet-overview}(a)), and is only employed during network training.

\subsubsection{Local Contrastive Module} \label{sec:3.4.1}

To increase target saliency and its feature distinctiveness to its neighboring background, we propose LCM, which uses a triple difference assessment to adaptively enhance the target regions while suppressing its neighboring regions. As shown in \cref{fig:cadd}(a), using the target ground-truth location, we extract the target region $\mathbf{R}_5\in\mathbb{R}^{C\times h\times w}$ and sample eight adjacent regions with matching area and aspect ratio to represent the background. We define this set of nine regions as $\mathbf{P}^{\text{in}}\in\mathbb{R}^{C\times H\times W}=\{\mathbf{R}_i\in\mathbb{R}^{C\times h\times w}\}_{i=1}^9$ where $H=3h$ and $W=3w$, and utilize it for local contrastive computation. Based on $\mathbf{P}^{\text{in}}$, we compute three distinct nine-dimensional feature vectors (i.e., average pooling, max pooling, and mean gradient pooling) defined as follows:

\vspace{-2em}
\begin{align}
    \mu_i&=\text{mean}(\mathbf{R}_{i}), \\
    m_i&=\text{max}(\mathbf{R}_{i}),\\
    \Delta_i&=\text{mean}\left(\sqrt{(\Delta_{x}\mathbf{R}_{i})^2+(\Delta_{y}\mathbf{R}_{i})^2}\right).
\end{align}

\vspace{-0.5em}
% \vspace{-0.8em}
\noindent where $\Delta_{x}\mathbf{R_{i}}$ and $\Delta_{y}\mathbf{R_{i}}$ denote the gradient on $x$ and $y$ axes of $\mathbf{R}_i$. 
% After obtaining these feature vectors, 
We then perform feature difference computation by subtracting each element with the target region to form $\bm{\mu}=\{\mu'_{i}\}^9_{i=1}$, $\bm{m}=\{m'_{i}\}^9_{i=1}$ and $\bm{\Delta}=\{\Delta'_{i}\}^9_{i=1}$ to indicate how different the neighboring regions are. For example, $\mu'_{i}\in\bm{\mu}$ can be computed as:
% elements in the average pooling difference vector $\bm{\mu}$ are calculated as:
\vspace{-0.5em}
\begin{align}
    \mu'_{i}=\left\{
        \begin{array}{ll}
            |\mu_{i}-\mu_{5}|,&\ \text{if}\ i\neq5 \\
            \mu_{5},&\ \text{else}
        \end{array}
    \right.
\end{align}
% \vspace{-0.5em}
% compared to the target region:
% \vspace{-0.5em}
% \begin{align*}
%     v_i'=v_i-v_5,\  i=1,...,9\ \text{and}\ v\in\{\mu,m,\Delta\},
% \end{align*}
\noindent where $\mu_5$ is for the target region. After obtaining $\bm{\mu}$, $\bm{m}$ and $\bm{\Delta}$, they are fed into a shared MLP with three linear layers, added together and passed through a softmax activation to form a regional weight vector. This vector will be region-wisely multiplied on the original $\mathbf{P}^{\text{in}}$, and forms the final output $\mathbf{P}^{\text{out}}=\{\mathbf{R}'_{i}\}^{9}_{i=1}$
% the nine regions $\{R_i\}_{i=1}^9$, and forms the final output $P^{out}=\{R_i'\}_{i=1}^9$ 
with target regions highlighted and surrounding regions suppressed. We eventually enforce a local contrastive loss between $\mathbf{P}^{\text{in}}$ and $\mathbf{P}^{\text{out}}$, which implicitly enforces our CCDNet to modulate the relation between the target regions and their surrounding regions:
% using $P^{in}$  and $P^{out}$:
\vspace{-0.8em}
{
\small
\begin{align}
    \bm{\mathcal{L}}_{\text{LCM}}&=-\frac{1}{HW}\sum_{i=1}^H\sum_{j=1}^W \Psi(\mathbf{P}_{ij}^{\text{in}},\mathbf{P}_{ij}^{\text{out}}), \\
    \Psi(\mathbf{P}_{ij}^{\text{in}},\mathbf{P}_{ij}^{\text{out}})&=\log 
\frac{||\mathbf{P}_{ij}^{\text{in}}-\mathbf{P}_{ij}^{\text{out}}||}{\text{max}(\mathbf{P}^{\text{in}},\mathbf{P}^{\text{out}})-\text{min}(\mathbf{P}^{\text{in}},\mathbf{P}^{\text{out}})},
\end{align}
}

\vspace{-0.5em}
% \vspace{-0.5em}
\noindent where 
% $M\ \text{and}\ N$ are the height and width of $P^{in}$ and $p^{out}$, 
$||\cdot||$ means the $l$-1 distance, $\text{max}(\mathbf{P}^{\text{in}},\mathbf{P}^{\text{out}})$ and $\text{min}(\mathbf{P}^{\text{in}},\mathbf{P}^{\text{out}})$ denote the maximal and minimal pixel value in $\mathbf{P}^{\text{in}}$ and $\mathbf{P}^{\text{out}}$. The goal of this loss is to maximize the feature difference between $\mathbf{P}^{\text{in}}$ and $\mathbf{P}^{\text{out}}$ with the target region enhanced and its surrounding regions suppressed, so the target can be more salient than its neighboring background, thus facilitating identifying the real targets from distractors.

\subsubsection{Global Contrastive Module} \label{sec:3.4.2}

To increase the network's discrimination of distractors in complex backgrounds, we design GCM (as in \cref{fig:cadd}(b)) to perform contrastive learning on real targets and their distractors. 
% One key challenge in this approach is to effectively define the negative samples. An intuitive way is to manually label the distractors in the images, but this is unequivocally laborious, and might introduce biases due to inconsistent standards of distractors. 
% In this work, 
We propose to adaptively choose negative samples by reusing detection results with relatively lower confidence scores. That is, for $M$ detection results $\mathbf{S}_{\text{det}}=\{\mathbf{x}_i\}_{i=1}^{M}$ in each iteration of network training, we assign positive and negative sample labels as:
\vspace{-0.5em}
\begin{align}
    \forall \mathbf{x}_i\in \mathbf{S}_{\text{det}},\ z_{\mathbf{x}_i}=\left\{
    \begin{array}{ll}
        z^+, & \text{if }\ \text{conf}(\mathbf{x}_i)>t_1, \\
        z^-, & \text{if }\ t_2<\text{conf}(\mathbf{x}_i)\le t_1, \\
        % z_{\text{pos.}}, & \text{if }\ x_{conf}>0.8, \\
        % x_{\text{neg.}}, & \text{if }\ 0.2<x_{conf}\le0.8, \\
        \varnothing, & \text{otherwise},
    \end{array}
    \right.
    \label{eq:gcm-threshold}
\end{align}

\vspace{-0.5em}
\noindent where $z^+$ and $z^-$ denote positive and negative sample labels, conf($\mathbf{x}$) means the confidence score of the detection result $\mathbf{x}$, and $t_{1}=0.8,t_{2}=0.2$ are confidence thresholds. Note that for the detection results of one image, we will select the highest score positive sample as the final positive sample, and the top $k=3$ highest score negative samples as the hard negative samples. With the samples achieved, we project the sample localizations on each output feature map from the backbone, extract their features and calibrate them into same-size vectors for contrastive learning. 
% The contrastive learning performs similarity computation on positive and negative samples, and
Our GCM employs a global contrastive loss to penalize the proposed network to meticulously differentiate the distractors in the backgrounds. The loss function is formulated as:
\vspace{-0.5em}
{
\small
\begin{align}
    % \bm{\mathcal{L}}_{\text{GCM}}=\sum_{a=1}^{N_{p}}\frac{-1}{N_p}\sum_{p=1}^{N_p}\log&\frac{\Phi(\mathbf{v}_a,\mathbf{v}_p)}{\sum_{p=1}^{N_p}\Phi(\mathbf{v}_a,\mathbf{v}_p)+\sum_{n=1}^{N_n}\Phi(\mathbf{v}_a,\mathbf{v}_n)}, \\
    \bm{\mathcal{L}}_{\text{GCM}}^{MT}=\sum_{p=1}^{N_{p}}\frac{-1}{N_p}\sum_{a=1}^{N_p}\log&\frac{\Phi(\mathbf{v}_a,\mathbf{v}_p)}{\sum\limits_{p'=1}^{N_p}\Phi(\mathbf{v}_a,\mathbf{v}_{p'})+\sum\limits_{n=1}^{N_n}\Phi(\mathbf{v}_a,\mathbf{v}_n)}, \\
    % \vspace{-0.5em}
    \Phi(\mathbf{v}_m,\mathbf{v}_n)&=\frac{\exp(\mathbf{v}_m\cdot \mathbf{v}_n)}{\tau},
\end{align}
}

\vspace{-0.5em}
\noindent where 
% $\mathbf{x}$ is the feature for similarity computation, 
$\mathbf{v}_a$, $\mathbf{v}_p$, and $\mathbf{v}_n$ are the anchor, the positive, and the negative samples, respectively. $N_p$ and $N_n$ denote the numbers of positive and negative samples. $\exp(\cdot)$ is the exponential cosine similarity, and $\tau=0.1$ is a temperature parameter. 
% \ul{Details of the design and the thresholds are in the supplement} \cref{sec:x-gcm}.

\subsection{Detection Head and Loss Function} \label{sec:3.5}

% We adopt the detection head as in \cite{fang2023dagnet,fang2023danet}. 
Our detection head composes of a classification and a regression heads, each of which has two 3x3 convolution layers. The loss function of our work is composed of two parts: the detection loss $\bm{\mathcal{L}}_{\text{det}}$, and the contrastive learning loss $\bm{\mathcal{L}}_{\text{LCM}}$ and $\bm{\mathcal{L}}_{\text{GCM}}$.
% $\bm{\mathcal{L}}_{\text{CaDD}}$. 
The detection loss consists of classification loss $\bm{\mathcal{L}}_{\text{cls}}$ (cross-entropy loss) and localization loss $\bm{\mathcal{L}}_{\text{loc}}$ (L2 loss).
% and the contrastive learning loss includes LCM loss $\bm{\mathcal{L}}_{\text{LCM}}$ and GCM loss $\bm{\mathcal{L}}_{\text{GCM}}$. 
The overall loss function is written as:
\vspace{-0.5em}
\begin{equation}
    \bm{\mathcal{L}}=\bm{\mathcal{L}}_{\text{cls}}+\bm{\mathcal{L}}_{\text{loc}}+\alpha\bm{\mathcal{L}}_{\text{LCM}}+\beta\bm{\mathcal{L}}_{\text{GCM}},
\end{equation}

\vspace{-0.5em}
\noindent where we set $\alpha=0.1$ and $\beta=0.05$ at training.
\section{Experiments}

\begin{table*}[t]
    \newcolumntype{C}[1]{>{\centering\arraybackslash}p{#1}}
    \fontsize{8.5}{10}\selectfont
    % \small
    \centering
    \caption{Quantitative results of our proposed method and other comparison methods. The best values are highlighted in bold fonts, and the second best are underlined. Our proposed method achieves the overall best results on different metrics and datasets.}
    \vspace{-1em}
    \begin{tabular}{c|c|c|c|c|c|c|c|c|c|c|c}
        \Xhline{1.2pt}
        \multirow{2}*{Category} & \multicolumn{2}{c|}{\multirow{2}*{Method}} & \multicolumn{3}{c}{IRSTD-1k \cite{zhang2022isnet}}\vline & \multicolumn{3}{c}{NUDT-SIRST \cite{li2022dnanet}}\vline & \multicolumn{3}{c}{NUAA-SIRST \cite{dai2021acm}} \\
        \cline{4-12}
         & \multicolumn{2}{c|}{} & P$_\uparrow$ & R$_\uparrow$ & F-1$_\uparrow$ & P$_\uparrow$ & R$_\uparrow$ & F-1$_\uparrow$ & P$_\uparrow$ & R$_\uparrow$ & F-1$_\uparrow$ \\
        \hline
        \multirow{3}*{\makecell[c]{General\\Detectors}} & RetinaNet  \cite{ross2017retinanet} & \textit{TPMAI'20} & 81.28 & 85.72 & 83.72 & 88.81 & 79.52 & 83.91 & 90.18 & 87.92 & 89.04 \\
         & Deformable DETR \cite{zhu2021deformabledetr} & \textit{ICLR'21} & 73.19 & 62.79 & 67.59 & 79.97 & 75.68 & 77.77 & 80.01 & 71.20 & 75.35 \\
         & FCOS \cite{tian2020fcos} & \textit{TPAMI'20} & 85.91 & 88.45 & 87.16 & 90.31 & 89.89 & 90.10 & 90.03 & 84.32 & 87.08 \\
        \hline
        \multirow{2}*{\makecell[c]{Camouflage\\Detectors}} & SINet \cite{fan2020sinet} & \textit{CVPR'20} & 80.18 & 84.28 & 82.17 & 76.52 & 81.37 & 78.87 & 86.94 & 88.20 & 87.57 \\
         & FEDER \cite{he2023feder} & \textit{CVPR'23} & 84.93 & 89.29 & 87.06 & 90.47 & 85.75 & 88.05 & 90.33 & 87.78 & 89.04 \\
        \hline
        \multirow{3}*{\makecell[c]{Traditional\\IRSTD\\Methods}} & PSTNN \cite{zhang2019infrared} & \textit{RS'19} & 61.17 & 58.88 & 60.01 & 66.67 & 64.09 & 65.35 & 63.48 & 72.51 & 67.70 \\
         & DNGM \cite{wu2020double} & \textit{GRSL'20} & 70.01 & 61.29 & 65.36 & 65.25 & 63.72 & 64.48 & 59.28 & 62.19 & 60.71 \\
         & TSLSTIPT \cite{sun2020infrared} & \textit{TGRS'20} & 69.69 & 62.88 & 66.11 & 65.15 & 67.78 & 66.44 & 75.28 & 77.31 & 76.28 \\
        \hline
        \multirow{6}*{\makecell[c]{DL-based\\IRSTD\\Methods}} & ACM \cite{dai2021acm} & \textit{WACV'21} & 89.98 & 88.68 & 89.32 & 90.37 & 91.25 & 90.81 & 89.72 & 90.13 & 89.92 \\
         & ISNet \cite{zhang2022isnet} & \textit{CVPR'22} & 88.69 & 83.87 & 86.21 & 90.08 & 88.78 & 89.43 & 92.81 & 88.84 & 90.78 \\
         & DAGNet \cite{fang2023dagnet} & \textit{TII'23} & 89.26 & 85.19 & 87.18 & 89.95 & 90.24 & 90.09 & 90.82 & 86.69 & 88.71 \\
         & MSHNet \cite{liu2024sls} & \textit{CVPR'24} & \underline{90.55} & \underline{89.43} & \underline{89.99} & \underline{91.88} & \textbf{92.84} & \underline{92.36} & \underline{93.49} & \underline{90.24} & \underline{91.84} \\
         & IRPruneDet \cite{zhang2024irprunedet} & \textit{AAAI'24} & 85.79 & 86.88 & 86.33 & 87.52 & 83.69 & 85.56 & 86.58 & 89.96 & 88.24 \\
         & \cellcolor{gray!80}\textbf{CCDNet (Ours)} & \cellcolor{gray!80}- & \cellcolor{gray!80}\textbf{92.89} & \cellcolor{gray!80}\textbf{90.23} & \cellcolor{gray!80}\textbf{91.54} & \cellcolor{gray!80}\textbf{92.08} & \cellcolor{gray!80}\underline{92.65} & \cellcolor{gray!80}\textbf{92.37} & \cellcolor{gray!80}\textbf{93.81} & \cellcolor{gray!80}\textbf{92.84} & \cellcolor{gray!80}\textbf{92.32} \\
        \Xhline{1.2pt}
    \end{tabular}
    \vspace{-1.6em}
    \label{tab:quantitative}
\end{table*}

\subsection{Experiment Setups}

\textbf{Datasets}: We conduct experiments on three publicly available IRSTD datasets: IRSTD-1k \cite{zhang2022isnet}, NUDT-SIRST \cite{li2022dnanet}, and NUAA-SIRST \cite{dai2021acm}. 
% We divide each dataset into training and testing sets by a ratio of 8:2, and perform data augmentation (image flippings) to complement the composition of each data. 
% We divide each dataset equally into 8:2 to ensure cross-dataset integrity, 
We adopt a consistent data split (train:test = 8:2 in IRSTD-1k \cite{zhang2022isnet}) to all datasets for a fair comparison, and perform data augmentation (image flippings) to complement the composition of each data. 
Each compared method in this paper is retrained on this augmented dataset for fair comparison, and is reported with their best performance with optimal settings. The dataset preparation detail can be found in \cref{sec:x-exp} in supplement.

\vspace{0.25em}
\noindent \textbf{Implementation Details}: The proposed method is implemented on one Nvidia V100 GPU and PyTorch framework. We adopt AdamW as the optimizer with an initial learning rate of 0.0001 and a weight decay of 0.0005. We train our proposed method for 150 epochs with a batch size of 12 using 32 hours in total.

\vspace{0.25em}
\noindent \textbf{Evaluation Metrics}: We use precision (P), recall (R), and F-1 score to assess the target-level performance. We also use model parameters (Params.), FLOPs and FPS to assess model complexity.
% A higher P score indicates less false alarm rate, and a higher R score indicates fewer missed detections. The F-1 score is a balanced metric of P and R, and a higher F-1 score suggests the detection method is more generally effective and robust.

\begin{table}[t]
    \centering
    \caption{Model complexity of IRSTD methods ranked by FPS.}
    \vspace{-0.5em}
    \small
    \begin{tabular}{c|c|c|c}
        \Xhline{1.2pt}
        Method & Params. (M)$_{\downarrow}$ & FLOPs (G)$_{\downarrow}$ & FPS$_{\uparrow}$ \\
        \hline
        % RetinaNet & 44.14 & 21.44 & 34.51 \\
        % Deformable DETR & 42.79 & 176.58 & 26.46 \\
        % FCOS & - & - & 22.83 \\
        ISNet \cite{zhang2022isnet} & 1.094 & 122.55 & 37.62 \\
        DAGNet \cite{fang2023dagnet} & 64.20 & 123.98 & 38.49 \\
        \rowcolor{gray!80} Ours & 65.88 & 121.54 & 39.76 \\
        ACM \cite{dai2021acm} & 0.520 & 8.385 & 42.74 \\
        IRPruneDet \cite{zhang2024irprunedet} & 0.180 & 0.938 & 51.98 \\
        MSHNet \cite{liu2024sls} & 4.07 & 33.51 & 69.54 \\
        \Xhline{1.2pt}
    \end{tabular}
    \label{tab:complex}
    \vspace{-1.5em}
\end{table}

\subsection{Comparisons with SOTA methods}

The comparison methods we choose include: the latest SOTA DL-based IRSTD methods (ACM \cite{dai2021acm}, ISNet \cite{zhang2022isnet}, DAGNet \cite{fang2023dagnet}, MSHNet \cite{liu2024sls}, IRPruneDet \cite{zhang2024irprunedet}); traditional IRSTD methods (PSTNN \cite{zhang2019infrared}, DNGM \cite{wu2020double}, and TSLSTIPT \cite{sun2020infrared}); general object detectors (RetinaNet \cite{ross2017retinanet}, Deformable DETR \cite{zhu2021deformabledetr}, and FCOS \cite{tian2020fcos}); camouflaged object detectors (SINet \cite{fan2020sinet} and FEDER \cite{he2023feder}).
% These methods are trained and tested on the same dataset above, and reported with their optimal performance after convergence.

% \vspace{0.25em}
\noindent \textbf{Quantitative Results}: The quantitative results of our proposed method and other comparison methods are given in \cref{tab:quantitative}. 
% As can be seen, our proposed method CIDNet surpasses the comparison methods in all metrics and all datasets.
% As in Tab.\ref{tab:quantitative}, 
% Generally speaking, DL-based IRSTD methods achieve the overall best detection results. 
General object detectors and camouflage object detectors
% achieve similarly decent detection performance, but are
achieve relatively inconsistent performance 
% across different datasets 
since they are not meticulously designed for IRSTD tasks.
% thus yielding poorer performance compared to DL-based IRSTD methods. 
For traditional IRSTD methods, their detection results are the poorest among all, because their models are
% are heavily based on handcrafted features that are 
not adaptive to complex detection challenges. For DL-based IRSTD methods, our proposed CCDNet surpasses almost all the others by an evident margin. For example, our CCDNet has obtained 1.55\% on IRSTD-1k, 0.01\% on NUDT-SIRST, and 1.48\% on NUAA-SIRST for F-1 metric higher than MSHNet (2nd place), verifying its effectiveness. 

Furthermore, we can see from \cref{tab:complex} that our CCDNet remains a decent real-time detection speed across other IRSTD methods during inference, suggesting that our method can detect more accurately with a real-time speed.

\begin{figure*}[t]
    \centering
    \includegraphics[width=\linewidth,keepaspectratio]{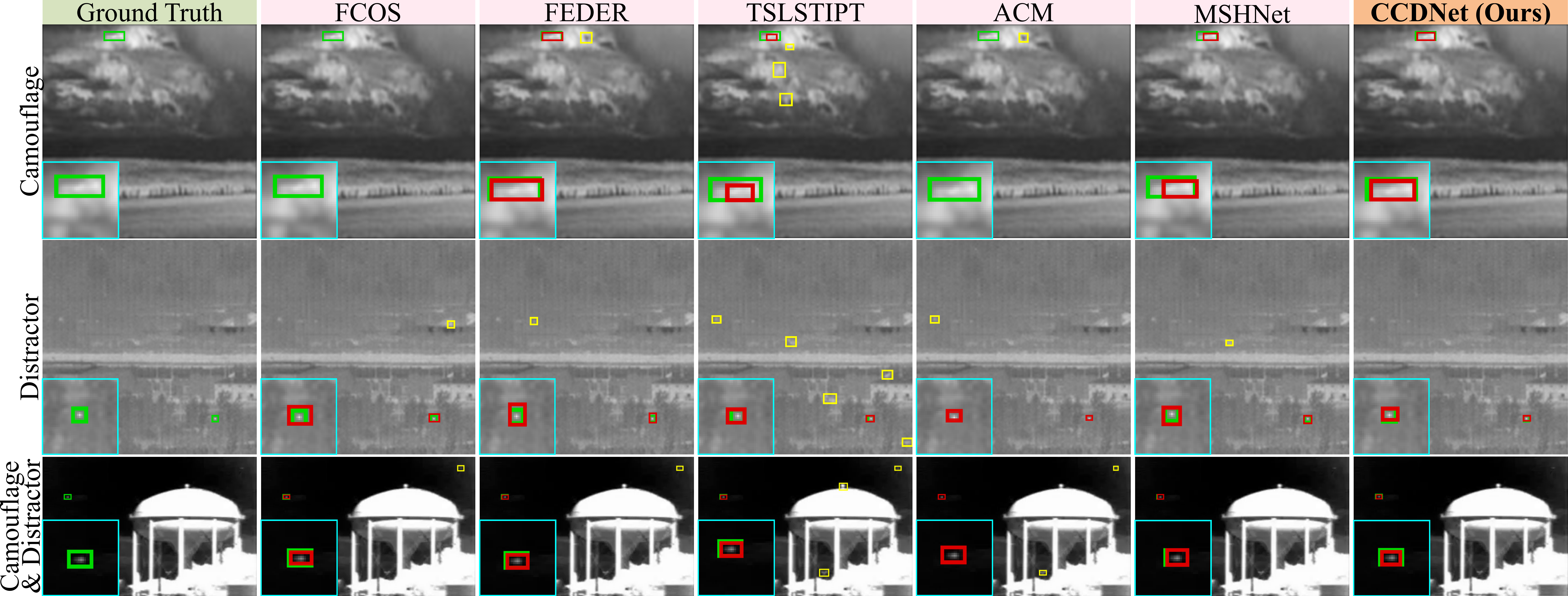}
    \vspace{-1.8em}
    \caption{Qualitative results of our proposed method and other comparison methods. For better visualization, we only pick at least one method from each category to showcase their detection performance. The \textcolor{red}{red}, \textcolor{green}{green}, and \textcolor{yellow}{yellow} boxes are ground truths, detection results, and false alarms. Image without any red box indicates a missed detection.
    % We circle the false alarms and missed detections to demonstrate the difference between the results more clearly. 
    We put close-ups for each detection result for better visual comparison. Qualitative comparisons of the other methods can be found in the supplement. Our CCDNet yields superior results with fewer false alamrs, fewer missed detections, and better overlapped bounding boxes.}
    
    % It can be seen that our CCDNet achieves superior detection results, with fewer false alarms and missed detections, and bounding boxes overlapping better with the ground truths.}
    \vspace{-1.5em}
    \label{fig:qualitative}
\end{figure*}

\vspace{0.25em}
\noindent \textbf{Qualitative Results}: The qualitative results are displayed in \cref{fig:qualitative}, where we show three DL-based IRSTD methods (ACM \cite{dai2021acm}, MSHNet \cite{liu2024sls} and ours), and one from each of the other method categories. The traditional IRSTD method TSLSTIPT \cite{sun2020infrared} has the poorest detection results with plenty of false alarms. Other comparison methods output better results but still suffer from misidentifying distractors as real targets.
% especially for FCOS \cite{tian2020fcos} and ACM \cite{dai2021acm} 
% where they fail to detect the target shown 
% in row 1. 
In contrast, our proposed method outputs the most accurate detections in complex detection scenes with camouflage and distractor interferences, and detection bounding boxes overlap with ground truths comparatively better than most other methods.
Qualitative results of more comparison methods can be found in \cref{sec:x-exp} in supplement. 
% This is because our proposed CCDNet can not only extract rich multi-level receptive field information through WMPs in the backbone and exploit them appropriately to represent the targets highlighted with TBSG and BOSE, but also discriminate irrelevant distractors from real targets by active contrastive learning via CaDD to reduce false alarms.

% The reason for the competitiveness of our CCDNet is two-fold. First, the proposed WMPs can extract target and background features with rich receptive fields, and the proposed ARFN fully exploits these features and remodels them to form a target-centric feature representation.
% % that facilitates accurate IRSTD. 
% This ensures the targets are highlighted for effective detection. Second, our proposed CaDD can adaptively strengthen the network's ability to discriminate distractors in the background via active local and global contrastive learning, which also improves detection performance.

\subsection{Ablation Study}

% We conduct ablation studies on the NUAA-SIRST \cite{dai2021acm} dataset, and the results have verified their complementarity to the baseline methods. 
We conduct the following ablation studies to verify the effectiveness of the proposed components. More ablations can be found in \cref{sec:x-exp} of the supplement.

\noindent \textbf{Impact of each proposed component.} We report the individual and incremental performance of the components of our method in \cref{tab:abl-0}. The baseline model in row 1 is the CCDNet with WMPs replaced by RepVGG block, and ARFN replaced by vanilla FPN. From rows 1 to 4, our proposed WMP, ARFN, and CaDD all achieve evident performance gain compared to the baseline, verifying their individual effectiveness. 
% WMP and CaDD achieve more performance gain than ARFN, which implies that WMP and CaDD are more influential to the model performance.
Additionally, from rows 1, 2, 5, and 6, our WMP, ARFN, and CADD can incrementally improve the performance, validating the design of our CCDNet.

\begin{table}[t]
    % \tiny
    % \fontsize{7}{8}\selectfont % 设置全局字体大小
    \small
    \centering
    \caption{Ablation study of each proposed component.}
    \vspace{-1em}
    \begin{tabular}{c|c|c|c|c|c|c}
        \Xhline{1.2pt}
        \# & WMP & ARFN & CaDD & P$_{\uparrow}$ & R$_{\uparrow}$ & F1$_{\uparrow}$ \\
        \hline
        1 & - & - & - & 87.68 & 88.09 & 87.89 \\
        % \hline
        2 & $\checkmark$ & - & - & 88.91 & 89.54 & 89.22 \\
        3 & - & $\checkmark$ & - & 88.33 & 89.25 & 88.79 \\
        4 & - & - & $\checkmark$ & 89.14 & 89.27 & 89.21 \\
        5 & $\checkmark$ & $\checkmark$ & - & 89.57 & 90.51 & 89.68 \\
        % \hline
        \rowcolor{gray!80} 6 & $\checkmark$ & $\checkmark$ & $\checkmark$ & \textbf{92.08} & \textbf{92.64} & \textbf{92.36} \\
        \Xhline{1.2pt}
    \end{tabular}
    \vspace{-1em}
    \label{tab:abl-0}
\end{table}

\begin{figure}[t]
    \centering
    \vspace{-0.3em}
    \includegraphics[width=\linewidth,keepaspectratio]{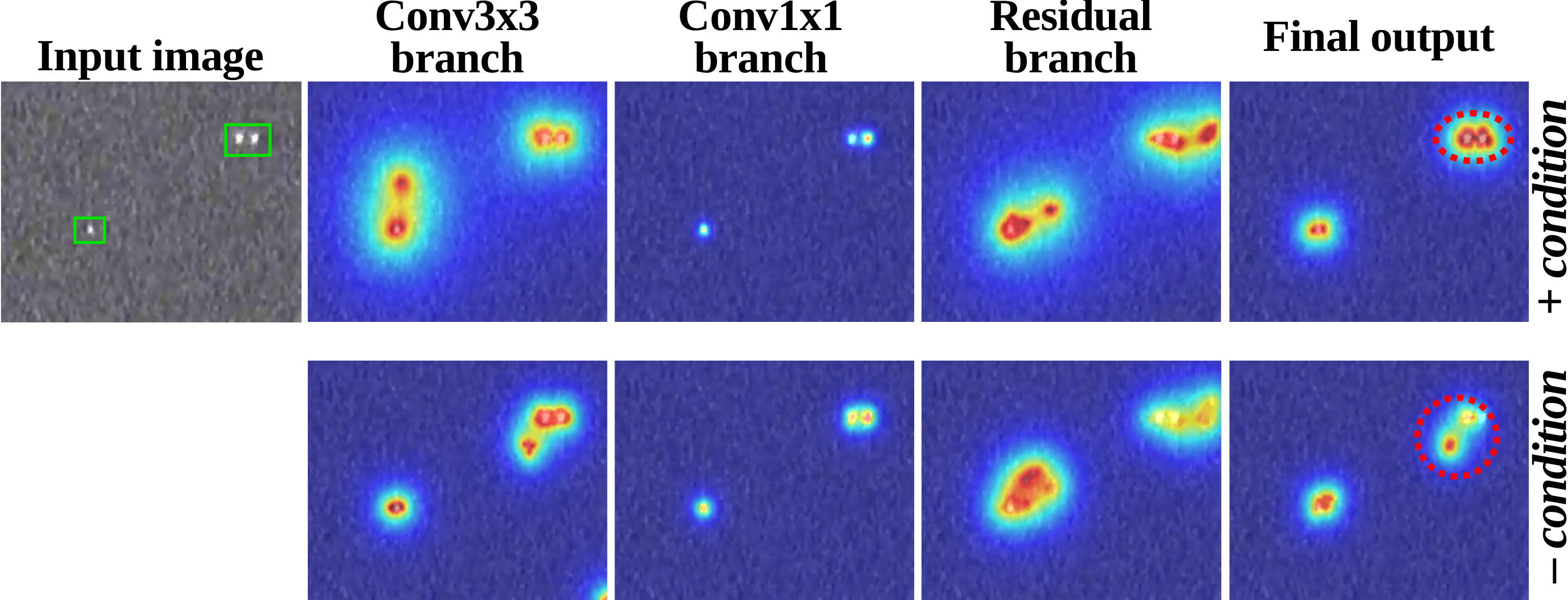}
    \vspace{-2em}
    \caption{Heatmaps for the outputs of last WMP from the stage 1 backbone. The results verify our WMP can accurately perceive the targets.}
    % With the multi-branch conditioning mechanism, our WMP can perceive more accurately on the target features.}
    \vspace{-1.8em}
    \label{fig:wmp-visual}
\end{figure}

% \vspace{0.25em}
\noindent \textbf{Effectiveness of the WMP}: We compare our WMP with various backbone blocks, as in \cref{tab:abl-wmp}, where WMP achieves the best detection performance among other convolutional blocks. 
% To further verify the adaptive self-conditioning mechanism in our WMP, 
We also plot the heatmaps for all three branches of the last WMP from our CCDNet's stage 1 backbone in \cref{fig:wmp-visual}. As can be seen, different branches in the WMP can capture different features in the image, and the adaptive self-conditioning mechanism can further effectively merge these features for an accurate representation of the targets, thus leading to a better performance.
% shows different feature awareness on three branches, and the final output with the conditioning mechanism yield the most accurate target feature perception. This shows that the adaptive conditioning mechanism can refine the multi-receptive field information for a better performance.
% our WMP’s multi-branch adaptive self-conditioning design allows it to effectively extract and utilize richer feature information, leading to superior results.
% Both ResNet block and bottleneck achieve relatively poorer results since their structure is single-branch. Although Inception and RepVGG have multi-branch structures, they can't adaptively control which branch is more dominant with rich useful feature information. 

% Thus, our WMP’s multi-branch adaptive self-conditioning design allows it to effectively extract and utilize richer feature information, leading to superior results.
% \vspace{-0.5em}
\begin{table}[t]
    \centering
    % \vspace{-0.5em}
    \caption{Ablation study on the proposed WMP.}
    \small
    \vspace{-1em}
    \begin{tabular}{c|c|c|c|c}
        \Xhline{1.2pt}
        \# & Backbone Type & P$_\uparrow$ & R$_\uparrow$ & F-1$_\uparrow$ \\
        \hline
        1 & ResNet Block \cite{he2016resnet} & 90.84 & 88.62 & 89.72 \\
        2 & ResNet Bottleneck \cite{he2016resnet} & 88.68 & 89.51 & 89.09 \\
        3 & Inception Block \cite{szegedy2015inception} & 88.92 & 90.17 & 89.54 \\
        4 & RepVGG Block \cite{ding2021repvgg} & 91.29 & 89.79 & 90.53 \\
        % \hline
        \rowcolor{gray!80} 5 & \textbf{WMP (Ours)} & \textbf{92.08} & \textbf{92.64} & \textbf{92.36} \\
        \Xhline{1.2pt}
    \end{tabular}
    \vspace{-1em}
    \label{tab:abl-wmp}
\end{table}

\vspace{-1em}

\vspace{1em}
\noindent \textbf{Effectiveness of TBSG and BOSE}: \cref{tab:abl-ff} presents ablation results on TBSG and BOSE. We evaluate several existing feature fusion mechanisms in CCDNet (rows 1-4), but none outperforms our proposed TBSG and BOSE. This is likely because alternative mechanisms consider using only target features while neglecting the distinct distributions of target and background features across backbone stages. Additionally, we assess TBSG and BOSE individually (rows 5-8), and both show significant performance improvement.
% the detection performance, with BOSE showing particular effectiveness by leveraging key target features to refine feature representations crucial for IRSTD tasks \cite{dai2021acm,dai2021alcnet,fang2023dagnet}. 

\begin{figure}[t]
    \centering
    \includegraphics[width=0.9\linewidth,keepaspectratio]{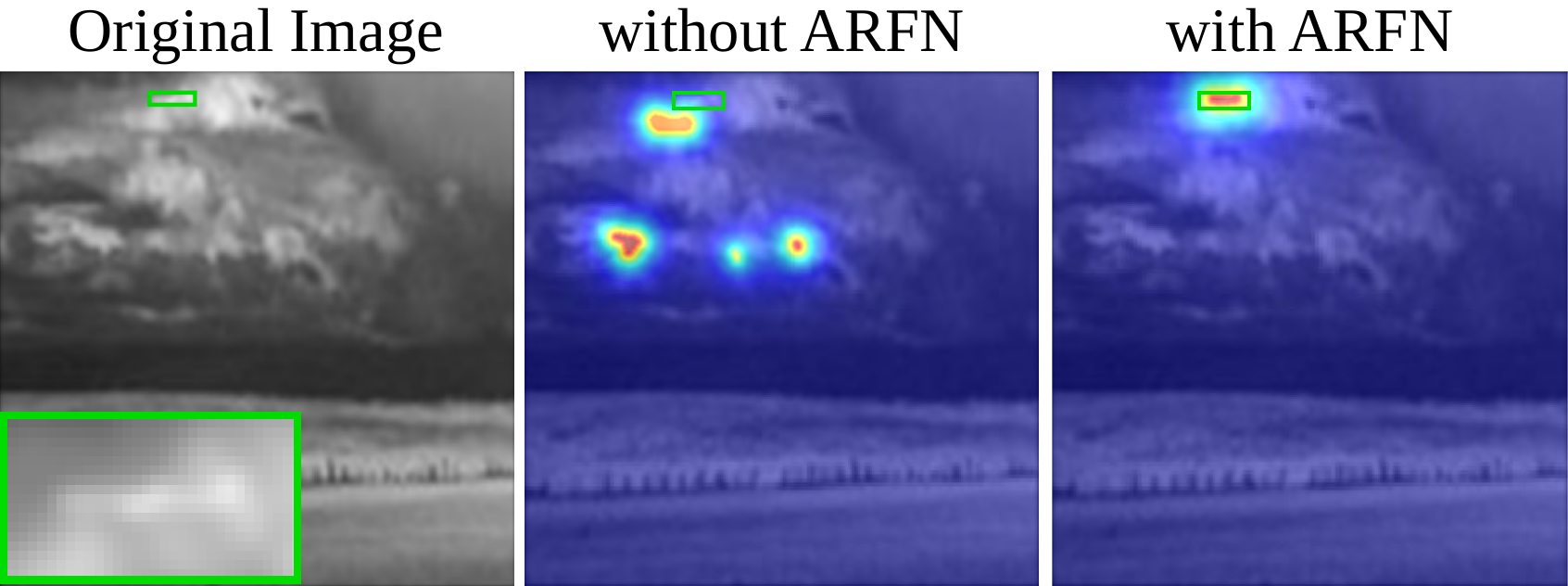}
    \vspace{-0.5em}
    \caption{Heatmap comparison of the first stage output feature map with and without using ARFN for camouflage detection. The green bounding box indicates the target region.}
    % and the results verify that ARFN can accurately perceive the camouflaged target.}
    \label{fig:arfn-visual}
\end{figure}

\begin{figure}[t]
    \centering
    \vspace{-1em}
    \includegraphics[width=\linewidth,keepaspectratio]{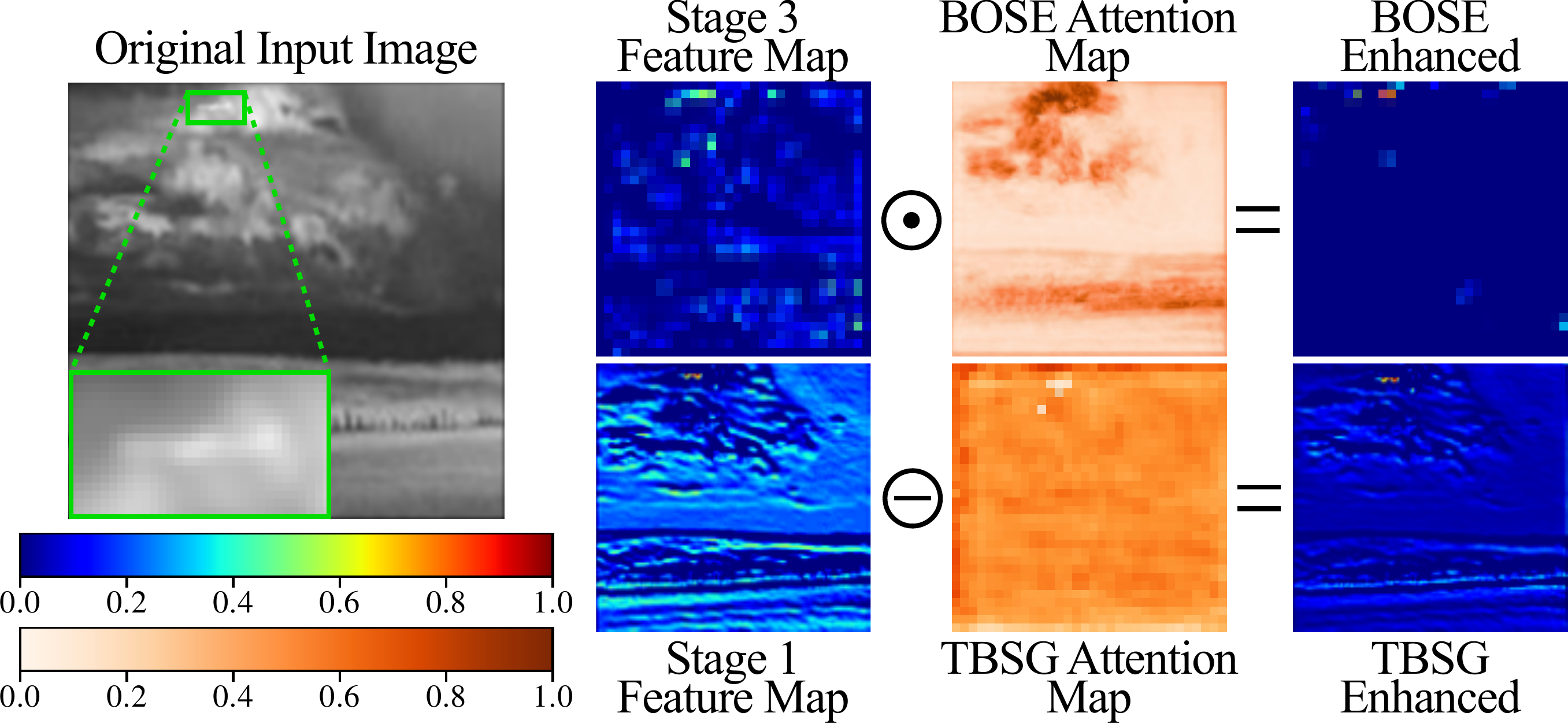}
    \vspace{-2em}
    \caption{Visual results of our proposed ARFN with BOSE and TBSG. This validates that ARFN can effectively highlight target features and suppress background features during feature fusion.}
    \vspace{-1em}
    \label{fig:bose-tbsg-visual}
\end{figure}

We provide the visual comparison of our CCDNet with and without the proposed ARFN in \cref{fig:arfn-visual}, where we illustarte the heatmaps for first stage output feature map. The target in the image is camouflaged into the mountain in the background, but our ARFN identifies its features successfully and precisely. Additionally, we visualize the feature maps along with the attention maps in BOSE and TBSG,
% to verify their ability to sense critical target and background features, 
as in \cref{fig:bose-tbsg-visual}. The visualizations show that both BOSE and TBSG can well perceive the target and the background features, and generate accurate attention maps to highlight the target regions and suppress the background.
% Thus, these validate the joint use of TBSG and BOSE in ARFN.

% \vspace{-0.5em}

% The joint use of TBSG and BOSE achieves the highest performance, validating their effectiveness in remodeling semantics and structures for a target-centric feature representation.
% \vspace{-0.5em}
\begin{table}[t]
    \centering
    % \vspace{0.5em}
    \caption{Ablation study on our proposed TBSG and BOSE.}
    \small
    \vspace{-1em}
    \begin{tabular}{c|c|c|c|c}
        \Xhline{1.2pt}
        \# & Fusion Type & P$_\uparrow$ & R$_\uparrow$ & F-1$_\uparrow$ \\
        \hline
        1 & FPN \cite{lin2017fpn} & 86.23 & 88.57 & 87.38 \\
        2 & PANet \cite{liu2018panet} & 88.96 & 89.08 & 89.02 \\
        3 & BiFPN \cite{tan2020efficientdet} & 85.25 & 86.48 & 85.86 \\
        4 & ACM \cite{dai2021acm} & 89.46 & 90.63 & 90.05 \\
        \hline
        5 & Ours w/o TBSG \& BOSE & 83.39 & 84.98 & 84.18 \\
        6 & Ours w/o TBSG & 86.12 & 85.38 & 85.75 \\
        7 & Ours w/o BOSE & 88.94 & 90.57 & 89.75 \\
        % \hline
        \rowcolor{gray!80} 8 & \textbf{Ours} & \textbf{92.08} & \textbf{92.64} & \textbf{92.36} \\
        \Xhline{1.2pt}
    \end{tabular}
    \vspace{-1.2em}
    \label{tab:abl-ff}
\end{table}

% \vspace{0.25em}
\noindent \textbf{Effectiveness of LCM and GCM in CaDD}: As shown in \cref{tab:abl-cadd}, incorporating either the LCM or GCM improves detection performance, with their combined use yielding the highest results. Compared to the baseline without any contrastive module, our CaDD improves P by 3.51\%, R by 2.43\%, and F-1 by 2.98\%. Notably, CaDD provides a greater boost in P than in R, likely because it enhances the model’s ability to distinguish distractors from true targets, and thus reduces false alarms.

\begin{figure}[t]
    \centering
    \includegraphics[width=\linewidth,keepaspectratio]{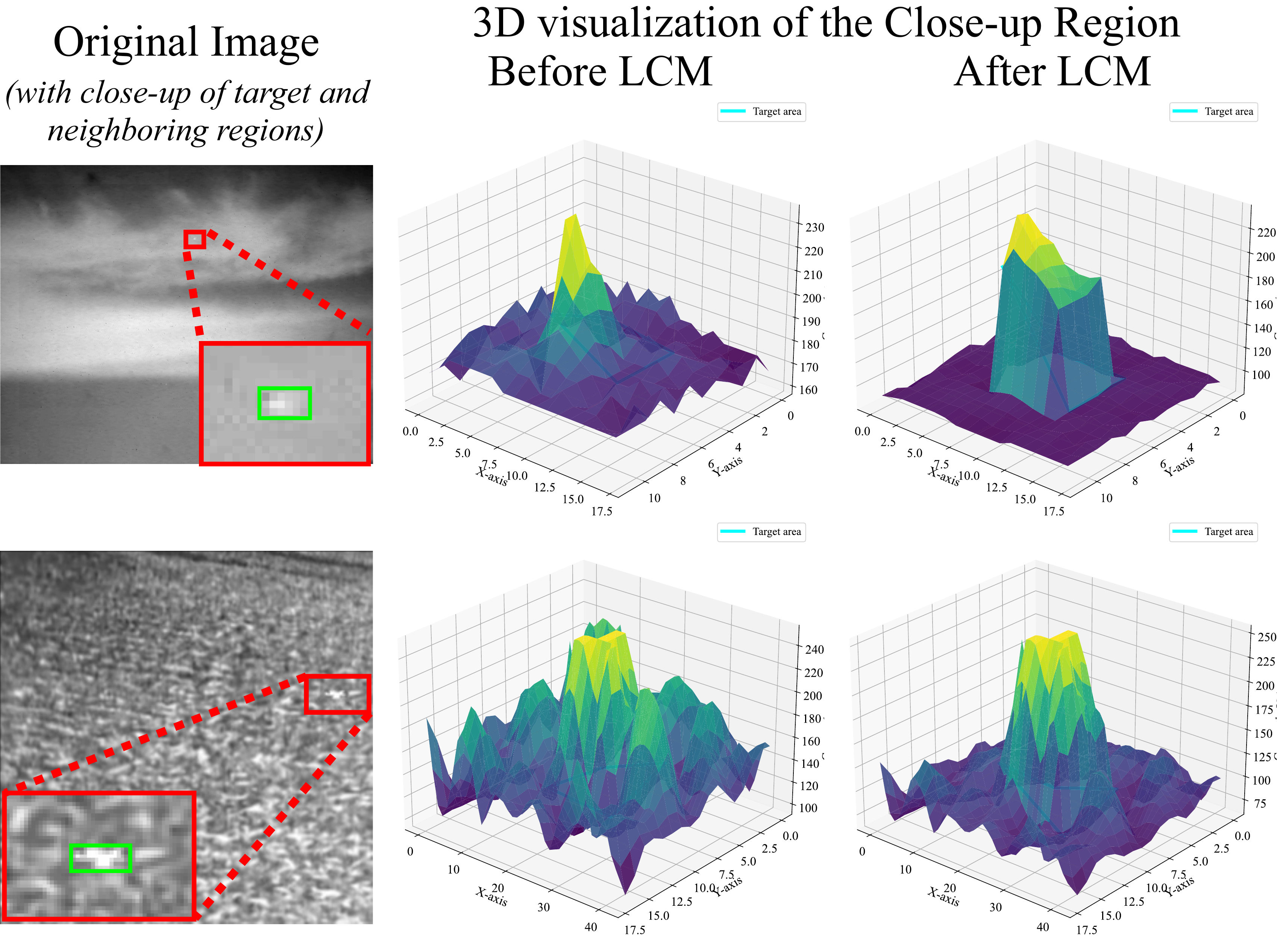}
    \vspace{-2.2em}
    \caption{3D Visualization of the impact of our proposed LCM.}
    % The results show that LCM is able to improve target saliency for better target discrimination against its neighboring surroundings.}
    \vspace{-1em}
    \label{fig:lcm-visual-main}
\end{figure}

\begin{figure}[t]
    \centering
    \includegraphics[width=\linewidth,keepaspectratio]{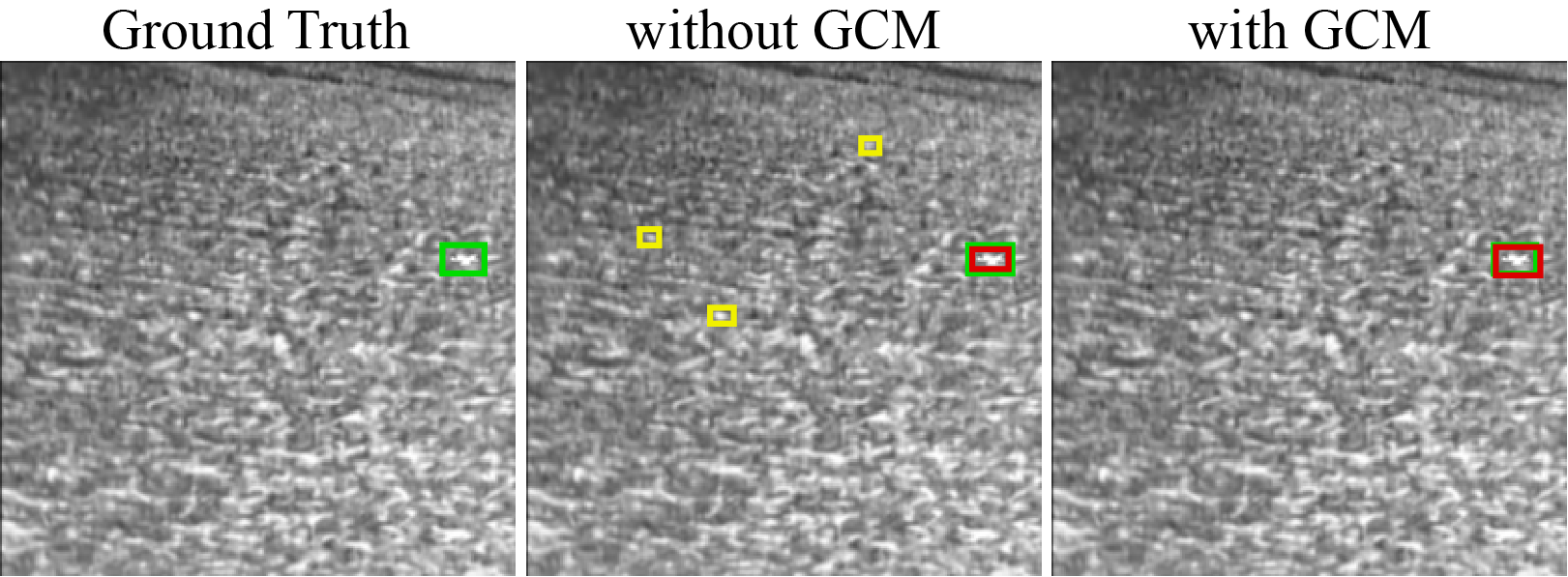}
    \vspace{-1.8em}
    \caption{Visualization of the effectiveness of the GCM. The \textcolor{red}{red}, \textcolor{green}{green}, and \textcolor{yellow}{yellow} boxes are ground truths, detection results, and false alarms.}
    \vspace{-1.5em}
    \label{fig:gcm-visual-main}
\end{figure}

We demonstrate two visualizations of the effect of our proposed LCM in \cref{fig:lcm-visual-main}. As can be seen from the original image, backgrounds surrounding the targets are relatively noisy, while the target features are not sufficiently discriminative. However, our proposed LCM is able to adaptively highlight the target regions and suppress the backgrounds effectively, thus making target features more identical.
% and further facilitating differentiating distractors from real targets in the image.
In addition, \cref{fig:gcm-visual-main} shows the difference in detection results of our CCDNet with and without GCM. It can be seen that, distractors are misidentified as real targets when our CCDNet is not employed with GCM, while using GCM leads to a precise detection with no misdetections. Both \cref{fig:lcm-visual-main} and \cref{fig:gcm-visual-main} verify the effectiveness of our proposed CaDD, which can improve the network ability to differentiate distractors and real targets in the image.
% Both \cref{fig:lcm-visual-main} and \cref{fig:gcm-visual-main} validate that our proposed CaDD, with LCM and GCM, is able to improve detection performance

% % As shown in Tab.\ref{tab:abl-cadd}, using either LCM or GCM can bring improvements to the detection performance, while employing both of them gives the best results. Compared to not using any contrastive module, our CaDD brings performance improvements of 3.51\% in P, 2.43 \% in R, and 2.98\% in F-1. It is noteworthy that using CaDD brings more improvements in metric P than R. This is because our CaDD can effectively strengthen the model's ability to discriminate distractors from the real targets, which brings down the false alarm rate.

\begin{table}[t]
    \centering
    \newcolumntype{C}[1]{>{\centering\arraybackslash}p{#1}}
    \caption{Ablation study on the proposed LCM and GCM.}
    \small
    \vspace{-1em}
    \begin{tabular}{c|C{1.1cm}|c|c|c|c}
        \Xhline{1.2pt}
        \multirow{2}*{\#} & \multicolumn{2}{c}{Contrastive Module}\vline & \multirow{2}*{P$_\uparrow$} & \multirow{2}*{R$_\uparrow$} & \multirow{2}*{F-1$_\uparrow$} \\
        \cline{2-3}
         & LCM & GCM & & & \\
        \hline
        1 & - & - & 88.57 & 90.21 & 89.38 \\
        2 & $\checkmark$ & - & 89.98 & 90.98 & 90.47 \\
        3 & - & $\checkmark$ & 90.79 & 91.03 & 90.91 \\
        % \hline
        \rowcolor{gray!80} 4 & $\checkmark$ & $\checkmark$ & \textbf{92.08} & \textbf{92.64} & \textbf{92.36} \\
        \Xhline{1.2pt}
    \end{tabular}
    \vspace{-1em}
    \label{tab:abl-cadd}
\end{table}

\vspace{-0.4em}
\section{Conclusion}
\vspace{-0.4em}

To address detecting camouflaged targets in complex backgrounds, and distinguishing distractors with similar features from true targets in IRSTD tasks, we introduce a novel IRSTD model CCDNet in this paper. For camouflaged targets, we develop a WMP-based backbone that capture multi-branch features with the proposed self-conditioning mechanism for accurate feature extraction. Additionally, we present ARFN, which uses feature refinement modules to build a target-centric feature representation, benefiting accurate target detection. 
To improve distractor discrimination, we propose CaDD
% We further propose CaDD with LCM and GCM, using two adaptive contrastive learning modules to improve distraction discrimination.
with LCM and GCM, adopting adaptive local and global contrastive learning to learn discriminative target features to reduce false alarms.
% The LCM enhances target saliency by re-modeling the relations between the target and neighboring regions, while the GCM iteratively selects distractor regions and enforces similarity computation with target regions to reinforce distractor distinction. 
Experimental results show that CCDNet outperforms various SOTA detection methods, achieving superior detection performance.

\clearpage
\setcounter{section}{5}
\maketitlesupplementary

% \section{Rationale}
% \label{sec:rationale}
% % 
% Having the supplementary compiled together with the main paper means that:
% % 
% \begin{itemize}
% \item The supplementary can back-reference sections of the main paper, for example, we can refer to \cref{sec:intro};
% \item The main paper can forward reference sub-sections within the supplementary explicitly (e.g. referring to a particular experiment); 
% \item When submitted to arXiv, the supplementary will already included at the end of the paper.
% \end{itemize}
% % 
% To split the supplementary pages from the main paper, you can use \href{https://support.apple.com/en-ca/guide/preview/prvw11793/mac#:~:text=Delete%20a%20page%20from%20a,or%20choose%20Edit%20%3E%20Delete).}{Preview (on macOS)}, \href{https://www.adobe.com/acrobat/how-to/delete-pages-from-pdf.html#:~:text=Choose%20%E2%80%9CTools%E2%80%9D%20%3E%20%E2%80%9COrganize,or%20pages%20from%20the%20file.}{Adobe Acrobat} (on all OSs), as well as \href{https://superuser.com/questions/517986/is-it-possible-to-delete-some-pages-of-a-pdf-document}{command line tools}.

\section*{Supplementary Material Overview}

% In this supplementary material, we present and discuss more details about our proposed method, and demonstrate more experiments to verify the effectiveness of our proposed method.
% % we present more implementation details about our proposed method in \cref{sec:x-implement}, and more experiment results in \cref{sec:x-exp}, including more quantitative and qualitative results, and more ablation studies. 
% A content list of this supplementary material is given as follows.

We have two main sections in our supplementary material. In \cref{sec:x-implement}, we provide some more architectural details of our proposed CCDNet (Camouflage-aware Counter-Distraction Network), including:

\begin{enumerate}[(1)]
    \setlength{\topsep}{0.6em}
    \setlength{\itemindent}{1em}
    \setlength{\itemsep}{0.15em}
    
    \item \cref{sec:x-wmp}: The transformation process of the proposed WMP (Weighted Multi-branch Perceptron) from training phase to inference phase.
    \item \cref{sec:x-arfn}: Motivations of our ARFN (Aggregation-and-Refinement Fusion Neck).
    \item \cref{sec:x-cadd}: Derivation of the CaDD (Contrastive-aided Distractor Discriminator) loss in network training, including LCM and GCM.
\end{enumerate}

In \cref{sec:x-exp}, we first clarify the preparation of our dataset and the result conversion in \cref{sec:x-data}, and then ablate the following to evaluate the validity of each of our designed modules in \cref{sec:x-ablation}:

\begin{enumerate}[(1)]
    \setlength{\topsep}{0.6em}
    \setlength{\itemindent}{1em}
    \setlength{\itemsep}{0.15em}
    
    % \item \cref{sec:x-condition}: The multi-branch self-conditioning mec-hanism introduced in our WMP.
    \item \cref{sec:x-refine}: Qualitative results of the effectiveness of our proposed ARFN, and the comparison of our ARFN employed with different feature refinement strategies.
    \item \cref{sec:x-lcm}: Impact of our proposed LCM (Local Contrastive Module) using different approaches of regional feature representation, and visualizations of the impact of our LCM.
    \item \cref{sec:x-gcm}: Discussion on the design of GCM (Global Contrastive Module), including using different global sampling strategies, different sample definition threshold pairs, and the comparison of sample selections for different global contrastive learning strategies.
\end{enumerate}

Finally, we present the qualitative detection results for the rest of the comparison methods in \cref{sec:x-restqua}.

\section{Implementation Details} \label{sec:x-implement}

% \textbf{Since I have not begun to compose the supplementary material, the following are just some random paragraphs for place holding.}

\subsection{Transformation of WMP} \label{sec:x-wmp}

\begin{figure}
    \centering
    \includegraphics[width=0.8\linewidth,keepaspectratio]{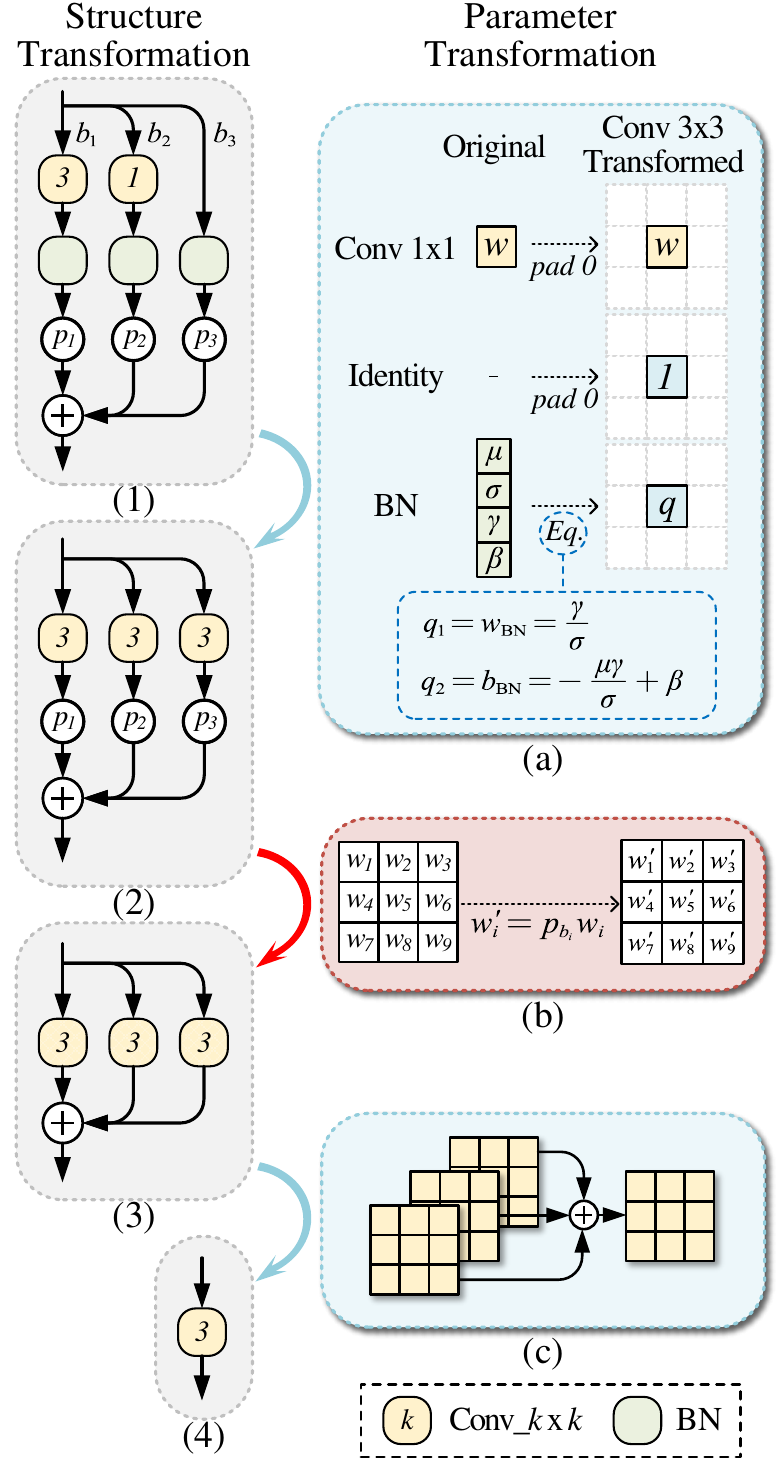}
    \caption{Structure transformation of our proposed WMP. States (1)-(4) show the structure transformations, and steps (a)-(c) demonstrate the parameter transformation. $\mu$, $\sigma$, $\gamma$, and $\beta$ denote the accumulated mean, standard deviation, and learned scaling factor and bias of the BN layer. The illustration of the learnable parameter integration is in (b). }
    % Further information about the rest of the transformation can be referred to \cite{x-ding2021repvgg}.}
    \vspace{-1em}
    \label{fig:x-wmp-trans}
\end{figure}

% Inspired by RepVGG \cite{x-ding2021repvgg}, our proposed WMP uses three branches to capture and aggregate multi-level features, which yield rich receptive field information favorable for accurately representing target and background features, thus facilitating our ARFN to fully exploit and model the relations between the targets and the backgrounds. To selectively refine the output of our proposed, we introduce an adaptive self-conditioning learnable parameter to regulate the importance of the features from each branch, aiming to filter out irrelevant information and reduce feature misalignment when fusing three branches.

% As introduced in \cref{sec:3.2}, 
Similar to \cite{x-ding2021repvgg}, our proposed WMP is different in the training and inference stages. During network training, the learnable parameter is directly imposed on the feature map of its corresponding branch, as the state (1) in \cref{fig:x-wmp-trans}. Therefore, the process of each branch in our WMP can be formulated as:
\begin{equation}
    \mathbf{F}_{b_i}^{\text{cond,(1)}}=\bm{\delta}_{b_i}(\mathbf{F}^{\text{in}})\cdot p_{b_i},\ i=1,2,3,
\end{equation}
where $\mathbf{F}^{\text{in}}$ is the input feature map, $\bm{\delta}_{b_i}(\mathbf{F}^{\text{in}})$ denotes the convolution and the BN operations on branch $b_i$, $p_{b_i}$ is the learnable contioning parameter on branch $b_i$, and $\mathbf{F}_{b_i}^{\text{cond}}$ is the conditioned feature map for branch $b_i$ in state (1).

During the inference stage, each branch will be transformed into a 3x3 convolution, and
% similar to \cite{x-ding2021repvgg}). 
a step-by-step illustration of the transformation process is shown in \cref{fig:x-wmp-trans}. For step (a) in \cref{fig:x-wmp-trans}, it is the same as in \cite{x-ding2021repvgg} where convolution and BN layers on each branch are transformed into a single 3x3 convolution. After that, our proposed adaptive conditioning mechanism is integrated into the 3x3 convolutional kernel without introducing any computational overhead. Define the conditioned feature map for branch $b_i$ of state (2) in \cref{fig:x-wmp-trans} as $\mathbf{F}_{b_i}^{\text{cond,(2)}}=\text{Conv}_{b_i}^{\text{(2)}}(\mathbf{F}^{\text{in}})\cdot p_{b_i}\in\mathbb{R}^{C\times H\times W}$, where $\text{Conv}_{b_i}^{\text{(2)}}(\cdot)=\mathbf{w}\in\mathbb{R}^{C_1\times C_2\times3\times3}\Leftrightarrow\bm{\delta}(\cdot)$ is the 3x3 convolution in state (2), where $C_1$ and $C_2$ are the input and output channel dimension. For each spatial location $l$ on $\mathbf{F}^{\text{in}}$, we can reformulate the convolution operation as:
% Define $F_{b_i}^{to}=\text{Conv}_{b_i}^{t}(F_{\text{input}})\cdot p_{b_i}\in\mathbb{R}^{C\times H\times W}$ and $F_{b_i}^{t}=\text{Conv}_{b_i}^{t}(F_{\text{input}})\in\mathbb{R}^{C\times H\times W}$, for each location $l$ on these feature maps, we can reformulate each branch of our proposed WMP:
\vspace{-0.5em}
\begin{equation}
    \begin{aligned}
    \mathbf{F}_{b_i}^{\text{cond,(2)}}(l)&=p_{b_i}\cdot\mathbf{w}\mathbf{F}^{\text{in}}(l) \\
    &=p_{b_i}\sum_k \mathbf{w}(k)\mathbf{F}^{\text{in}}(l+k) \\
    &=\sum_k\left( p_{b_i}\mathbf{w}(k) \right)\mathbf{F}^{\text{in}}(l+k) \\
    &=\sum_k \mathbf{w}'(k)\mathbf{F}^{\text{in}}(l+k) \\
    &=\mathbf{w}'\mathbf{F}^{\text{in}}(l) \\
    \end{aligned}
    \label{eq:wmp_infer}
\end{equation}

\vspace{-0.5em}
\noindent where $k\in[(-1,-1),(-1,0),...,(0,0),...,(1,0),(1,1)]$ is the kernel offset. Therefore, we can define $\text{Conv}_{b_i}^{\text{(3)}}(\cdot)=\mathbf{w}'\in\mathbb{R}^{C_1\times C_2\times3\times3}\Leftrightarrow\text{Conv}_{b_i}^{\text{(2)}}(\cdot)\cdot p_{b_i}=\mathbf{w}\cdot p_{b_i}$, thus fusing the learnable conditioning parameter into the 3x3 convolutional kernel, realizing step (b) in \cref{fig:x-wmp-trans}. Finally, we merge three 3x3 convolutional kernels together and form one single 3x3 convolution in step (c), thus achieving state (4).
% $w$ is the original 3x3 convolutional kernel weight, and $w'=p_{b_i}w$. 

We can observe from \cref{eq:wmp_infer} that the learnable parameter can be integrated into the convolutional kernel weight only involving a simple linear computation. However, this conditioning mechanism enables the network to adaptively select important target features for fusion instead of treating each branch equally, thus is beneficial for improving detection performance.

% We observe in Eq. \cref{eq:wmp_infer} that, the learnable parameter can be integrated into the convolutional kernel weight, and thus won't affect the branch transformation, bringing no extra computation on the fly. As shown in \textbf{\cref{fig:x-wmp-trans}}, from (2) to (3), the parameter transformation process (b) is the integration of the learnable parameter into the convolutional kernel, which only involves a simple linear computation to the kernel weight. This means our proposed WMP can not only transform exactly like the RepVGG block at the inference stage to improve the network's efficiency, but also maintain its ability to condition the fusion for each branch that benefits the network's detection accuracy.
% , achieving a delicate balance.

% \subsection{Breaking down our proposed ARFN} \label{sec:x-arfn}
\subsection{Motivation of our proposed ARFN} \label{sec:x-arfn}

We introduce ARFN in \cref{sec:3.3} to better exploit target and background features within our network, which ultimately benefits the detection performance of infrared small targets. In this section, we discuss the initial thought of our ARFN, and the feasibility of using background features in related vision tasks.

\textbf{Challenge of exploiting usable infrared small target features}. 
% In IRSTD tasks, the weakness of target features (e.g., relatively small target size, ambiguous contours, and low contrast to its background) has always been one key challenge to the IRSTD performance. 
Due to their limited spatial dimensionality, infrared small targets contain very few semantics and are quite liable to the network's downsampling operations as they will collapse their spatial feature integrity \cite{x-dai2021acm}, resulting in even sparser target features in the deep network layers that are difficult to exploit. That is to say, usable target features usually only exist at the shallow stages of the backbone, while the deep stages contain rich features of the backgrounds instead.

\textbf{Limitations for current IRSTD methods}. Many IRSTD methods pay attention to this issue. For example, DNA-Net \cite{x-li2022dnanet} employs dense connections between shallow and deep layers to ensure crucial features at shallow layers can be incorporated into deep layers, and DAGNet \cite{x-fang2023dagnet} proposes three different attention mechanisms to enhance target features to avoid their disappearance. However, sometimes complex background features 
% with relatively higher illuminance and more complicated details 
are usually \textbf{much more dominant} to target features, making the targets especially hard to identify in a direct way for many current IRSTD methods.
% posing a severe imbalance of feature distribution on the image and making target features difficult to excavate. 
% Many current IRSTD methods often lack consideration of this situation and thus yield suboptimal detection performance.

\textbf{Feasibility in adopting background features for target representation learning}. To address this issue, we propose to exploit rich background features to help direct the detection of the targets. We discover some previous works with similar initial thoughts:

\begin{figure}[t]
    \centering
    \includegraphics[width=\linewidth,keepaspectratio]{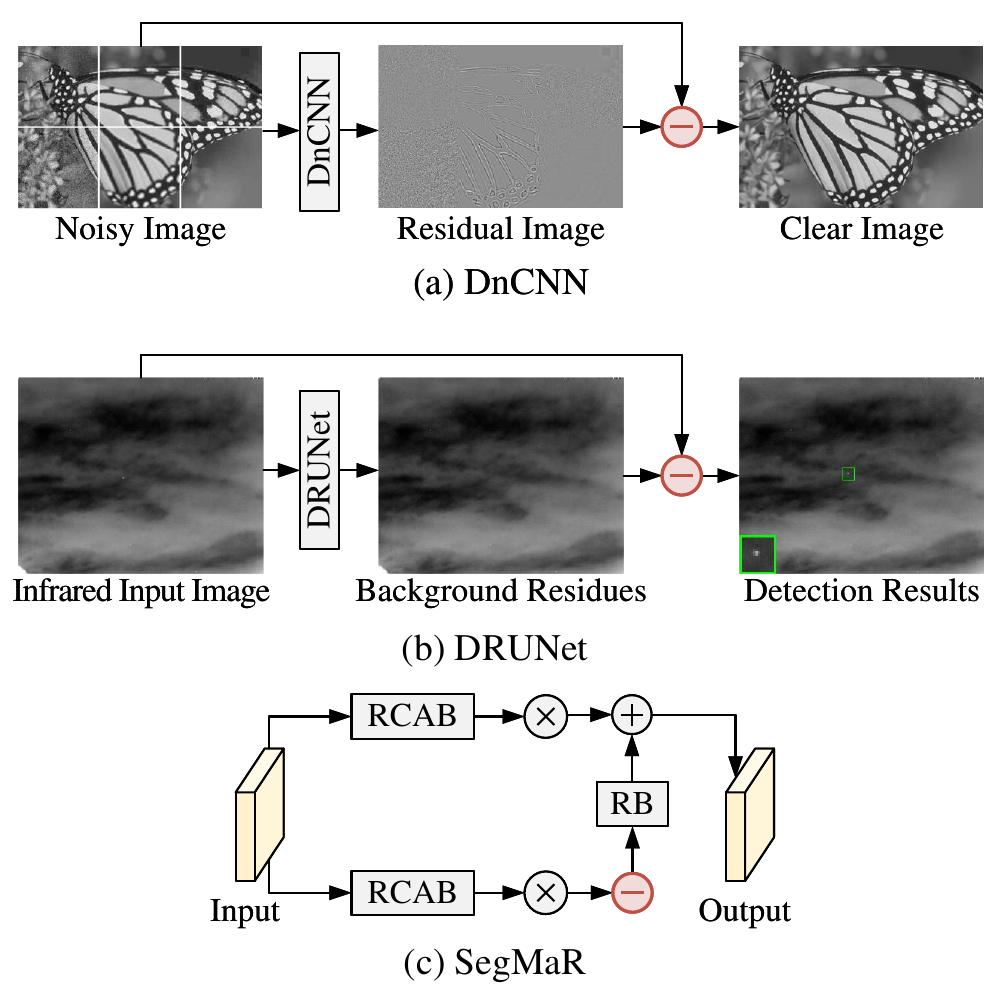}
    % \vspace{-2em}
    \caption{Related works using background features to guide vision tasks. Our ARFN studies this idea and propose to use background features to form a target-centric feature representation favorable for detection.}
    \vspace{-1em}
    \label{fig:arfn-related}
\end{figure}

\begin{enumerate}

    \item Proposed in 2017, DnCNN \cite{x-zhang2017dncnn} aims to solve the image denoising problem by predicting the residual image (i.e., the noises) instead of the clean image, as shown in \textbf{\cref{fig:arfn-related} (a)}.
    \item In IRSTD tasks, DRUNet \cite{x-fang2021drunet} takes this characteristic into account, and proposes to exploit the background features to reconstruct the target regions for detection, as shown in \textbf{\cref{fig:arfn-related} (b)}.
    \item As in SegMaR \cite{x-jia2022segmar}, it uses element-wise subtraction to reverse the background feature and element-wisely add it to augment the foreground feature, which regards background features as implicit inverse visual signals to highlight the camouflaged targets, as shown in \textbf{\cref{fig:arfn-related} (c)}.
\end{enumerate}

% All these works confirm the feasibility of taking advantage of background features for IRSTD tasks, and 
In conclusion, we base on this initiative to derive our proposed ARFN, which uses the proposed TBSG and BOSE to form a target-centric feature representation that facilitates improving infrared target detection performance.

\subsection{Derivation of CaDD Loss} \label{sec:x-cadd}

% In our main paper \cref{sec:3.4}, we simply demonstrate how our proposed CaDD works with only one target in the image, using only a single-level output feature map from the backbone, and the batch size is only 1. In this part, we will extend this to multiple targets, multiple levels, and multiple batches.

% In our main paper \cref{sec:3.4}, we introduce our proposed CaDD with LCM and GCM with a simple demonstration of how they work with one image. 

% \textbf{Multiple targets}. 
% Assume there are $M\ge1$ targets in an input image.
% with their ground truth labels $t_m=(x_m,y_m,w_x,h_m)$ representing their top-left coordinates, width, and height. 
% In our main paper \cref{sec:3.4}, we introduce our proposed CaDD with LCM and GCM with a simple demonstration.

\subsubsection{LCM Loss}

In LCM, since there might be more than 1 target in the image, we repeat the loss calculation for each target described in \textbf{\cref{sec:3.4.1}}, and obtain a mean loss for all targets in the image. Therefore, the LCM loss for this multi-target situation can be written as:
\vspace{-0.5em}
\begin{equation}
    \bm{\mathcal{L}}_{\text{LCM}}^{\text{MT}}=\frac{1}{M}\sum_{m=1}^{M}\bm{\mathcal{L}}_{\text{LCM}}(\mathbf{P}_{m}^{\text{in}},\mathbf{P}_{m}^{\text{out}}),
\end{equation}

\vspace{-0.5em}
\noindent where $\mathbf{P}_{m}^{\text{in}}$ and $\mathbf{P}_{m}^{\text{out}}$ are the $\mathbf{P}^{\text{in}}$ and $\mathbf{P}^{\text{out}}$ for the $m$-th target in the image, $M$ is the number of targets in the image, and $\text{MT}$ stands for multiple targets. It is noteworthy that there might be cases where a neighboring region overlaps with another target region. We deal with this by enforcing NMS between all neighboring regions and target regions with a threshold of 0.3, and fixing the loss to 0 for those overly overlapped background regions to avoid treating targets as backgrounds. 

We extend this loss for different levels of output feature maps from the backbone, as well as multiple training batches. Define $L$ as the number of output feature maps to be used in LCM, and $B$ is the batch size, we could have the extended LCM loss as the following:
\vspace{-0.5em}
{
\small
\begin{align}
    \bm{\mathcal{L}}_{\text{LCM},l}^{\text{MT}}&=\frac{1}{m_b}\sum_{m=1}^{m_b}\mathcal{L}_{\text{LCM}}(P_{m}^{in},P_{m}^{out}),m_b\in M_B  \label{eq:lcm-loss1} \\
    \bm{\mathcal{L}}_{\text{LCM},b}^{\text{ML}}&=\frac{1}{L}\sum_{l=1}^{L}\bm{\mathcal{L}}_{\text{LCM},l}^{\text{MT}},L=4 \label{eq:lcm-loss2} \\
    \bm{\mathcal{L}}_{\text{LCM}}^{\text{ext}}&=\bm{\mathcal{L}}_{\text{LCM}}^{\text{MB}}=\frac{1}{B}\sum_{b=1}^{B}\bm{\mathcal{L}}_{LCM,b}^{\text{ML}} \label{eq:lcm-loss3}
\end{align}
}

\vspace{-0.5em}
\noindent where ML, MB and ext stand for multiple levels, multiple batches, and extended function, respectively. \cref{eq:lcm-loss1}, \cref{eq:lcm-loss2} and \cref{eq:lcm-loss3} show that the LCM loss of each level and each batch are calculated individually and averaged by numbers of targets, levels, and batches, respectively. 

\subsubsection{GCM Loss}

In GCM, following the sample definition in \textbf{\cref{sec:3.4.2}}, we can have the positive and negative sample set in the image as $\mathbf{S}_{\text{pos}}=\{\mathbf{x}_i|z_{\mathbf{x}_i}=z^+ \text{ and } \mathbf{x}_i\in \mathbf{S}_{\text{det}}\}$ , $\mathbf{S}_{\text{neg}}=\{\mathbf{x}_i|z_{\mathbf{x}_i}=z^- \text{ and } \mathbf{x}_i\in \mathbf{S}_{\text{det}}\}$, and $\mathbf{S}=\mathbf{S}_{\text{pos}}\cup\mathbf{S}_{\text{neg}}$. Based on this, we can apply contrastive learning on samples in these two sets using the GCM loss function in \textbf{\cref{sec:3.4.2}}, with the only difference being that the numbers of samples are different (but the ratio is still kept as 1:3). Therefore, the GCM loss function for multiple targets can be written as:
{
\small
\begin{align}
    \bm{\mathcal{L}}_{\text{GCM}}^{MT}=\sum_{p=1}^{N_{p}}\frac{-1}{N_p}\sum_{a=1}^{N_p}\log&\frac{\Phi(\mathbf{v}_a,\mathbf{v}_p)}{\sum_{p'=1}^{N_p}\Phi(\mathbf{v}_a,\mathbf{v}_{p'})+\sum_{n=1}^{N_n}\Phi(\mathbf{v}_a,\mathbf{v}_n)}, \\
    \Phi(\mathbf{v}_m,\mathbf{v}_n)&=\frac{\exp(\mathbf{v}_m\cdot \mathbf{v}_n)}{\tau},
\end{align}
}

\noindent where $N_p=|\mathbf{S}_{\text{pos}}|$, $N_n=|\mathbf{S}_{\text{neg}}|$, $\mathbf{v}_a,\mathbf{v}_p\in \mathbf{S}_{\text{pos}}$, $\mathbf{v}_n\in S_{\text{neg}}$, and $\tau=0.1$ is a temperature hyperparamter. Note that at the beginning of the network training, there exists a chance that positive and negative sample definitions might not be satisfied due to low confidence scores. In this situation, we use ground truths as positive samples, and the top-3 detection results as the corresponding negative samples, until the sample definitions can be stably satisfied. 

The GCM loss can also be extended to multiple levels and batches, but we stress that global contrastive loss should be applied between target and distractor regions on the same level, since cross-level feature similarity computation might lead to semantic mismatch. Therefore, the extended GCM loss is written as:
\vspace{-0.5em}
\begin{align}
    \bm{\mathcal{L}}_{\text{GCM}}^{\text{MB}}=\frac{1}{L}\sum_{l=1}^{L}\frac{1}{B}\sum_{b=1}^B\bm{\mathcal{L}}_{\text{GCM},l,b}^{\text{MT}}
\end{align}

\vspace{-0.5em}
It is noteworthy that, our GCM loss is different from InfoNCE loss regularly used in self-supervised contrastive learning \cite{x-chen2020simple,x-he2020momentum,x-caron2021dino}, since we use already labeled data to further define pairs of positive samples (real targets) and negative samples (objects with similar features) related to IRSTD tasks. However, compared to the original supervised contrastive learning loss introduced in \cite{x-Khosla2020supervised}, which calculates contrastive loss for all samples in the batch, our GCM only computes that for positive samples in the batch.
% , because we want our network to only focus on learning fine-grained target features. 
This is because including contrastive loss calculations of negative samples may induce our network to incorrectly pay attention to learning distractor features, contradicting the purpose of GCM, which is to make our proposed network only focus on learning fine-grained target features. Therefore, we choose to compute contrastive loss for positive samples only.

\section{More Experiment Results} \label{sec:x-exp}

\subsection{Dataset and Result Preparation} \label{sec:x-data}

\vspace{-0.8em}
\begin{figure}[!htbp]
    \centering
    \includegraphics[width=\linewidth]{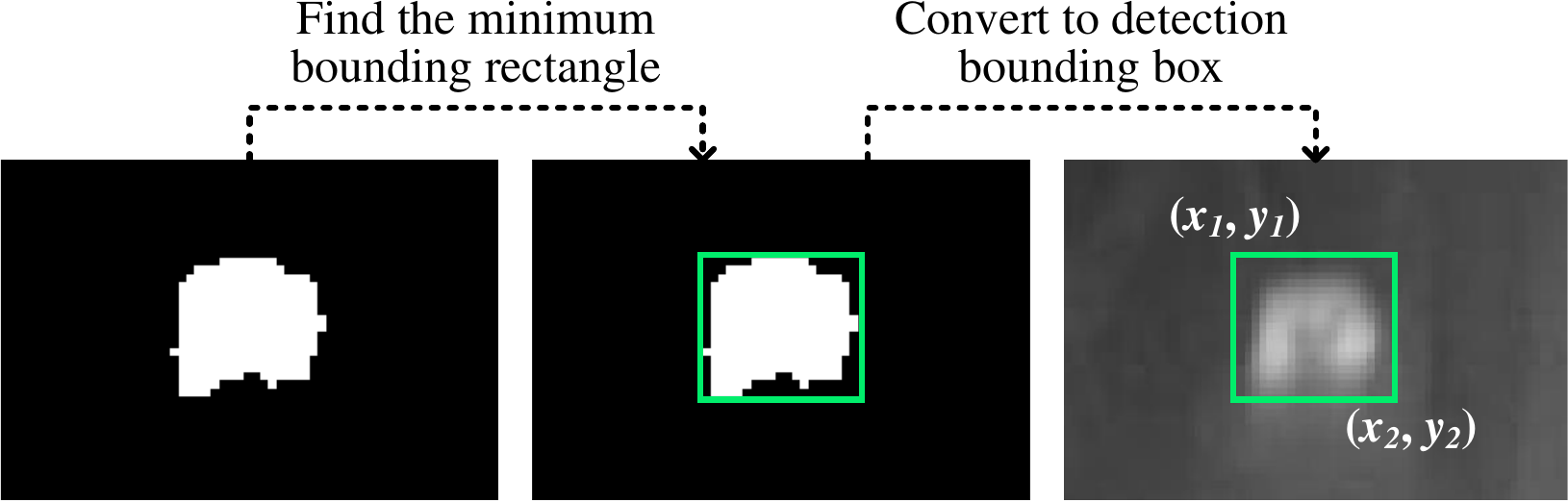}
    \vspace{-1.5em}
    \caption{Illustration of the data conversion process.}
    \vspace{-0.5em}
    \label{fig:conversion}
\end{figure}

\textbf{Segmentation dataset conversion}. Since our proposed method is for object detection while the datasets we use are for object segmentation, we modified the labels of the datasets. We load every mask image in the dataset and find the minimum bounding rectangle (as in \textbf{\cref{fig:conversion}}) for every target as its bounding box label for object detection. We then convert the box information in every image into a common object detection label format, e.g., VOC2007 XML files. We use detection datasets for object detection methods, and segmentation datasets for segmentation methods.
% and eventually convert the whole datasets into a form of VOC2007 datasets. 
% We train and test our proposed method on these converted datasets, and use the original segmentation datasets to train other segmentation methods. We also convert the VOC2007-structured datasets into COCO-structured ones for some specific detection methods.

\textbf{Detection results conversion}. Outputs of most segmentation methods are binary masked images with pixels of 255 (or 1) for target regions and 0 for backgrounds. We also find the minimum bounding rectangle for the target regions to yield the detection results for metric calculations. For some segmentation methods that output masked images with float data (e.g., floating numbers between 0 and 1), we use the OSTU's method to threshold the masked images, obtain the binary ones, and then convert them to detection results following the same procedure.

\subsection{More Ablation Studies} \label{sec:x-ablation}

% \subsubsection{Multi-branch Self-conditioning in WMP} \label{sec:x-condition}

\subsubsection{Impact of our proposed ARFN} \label{sec:x-refine}

\begin{figure*}[t]
    \centering
    \includegraphics[width=0.9\linewidth,keepaspectratio]{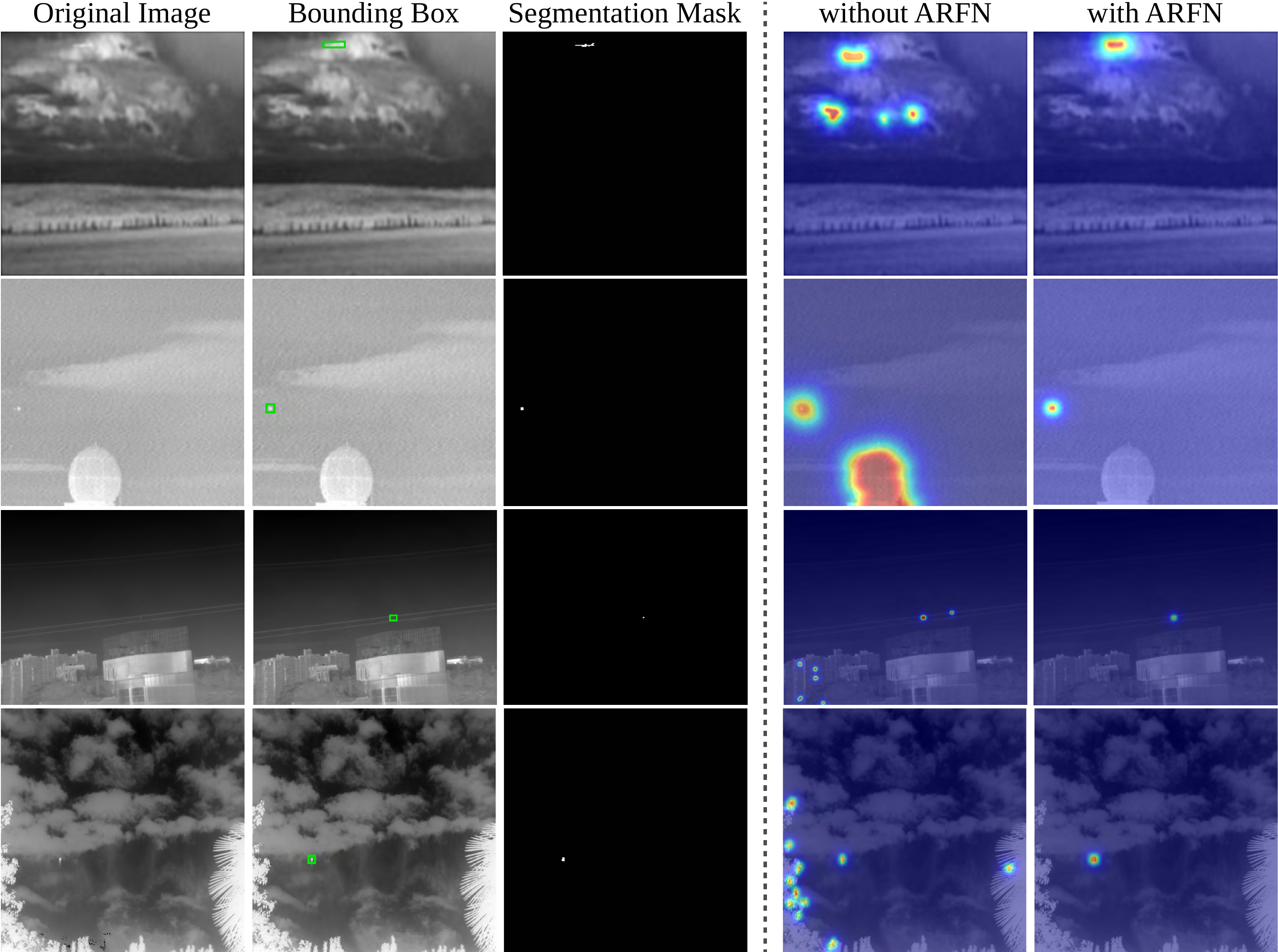}
    \vspace{-0.5em}
    \caption{Network visual heatmap comparisons between our CCDNet without and with the use of ARFN. The results suggest that our proposed ARFN can facilitate our CCDNet focusing more on the real targets and less on the backgrounds. Details can be found in \cref{par:fig8-epl}.}
    % line \hyperlink{line-983}{983}-1005.}
    \vspace{-0.5em}
    \label{fig:arfn-cam}
\end{figure*}

\hypertarget{line-983}{} \ \ \ \textbf{Visual results of the effectiveness of the proposed ARFN}. We provide the network visual heatmaps in \textbf{\cref{fig:arfn-cam}} to demonstrate the effectiveness of our proposed ARFN. For better visualization, we opt for the first-stage output feature map from the ARFN since it has the highest resolution and can better visualize details. The heatmaps generated without ARFN are obtained directly from the first stage of the backbone. It can be seen that our CCDNet can focus mainly on the target features when ARFN is integrated, which contributes to the accurate detection of them. \label{par:fig8-epl}

However, when CCDNet is not employed, the network tends to pay attention to irrelevant background features with similar features. In \textbf{\cref{fig:arfn-cam}} row 1, the target is almost camouflaged in its surrounding backgrounds, and the network misfocuses on the backgrounds to a great extent. In row 2, the network is severely misled by the background dominance. Although rows 3 and 4 don't have as much concentration on the backgrounds, they still attend to multiple distractor regions that can lead to detection performance degradation. Our proposed ARFN utilizes rich structures from shallow stages and abundant semantics from the backgrounds to adaptively form a target-centric feature representation, thus improving the detection performance in IRSTD tasks.

\hypertarget{line-1006}{} \textbf{Performance of using different feature refinement}. In \textbf{\cref{tab:arfn-fr}}, we present the performance of our TBSG and BOSE employed with or without any feature refinement mechanism. We investigate various attention mechanisms in our proposed modules, and discover that the channel refinement and dynamic spatial refinement modules we use can improve detection performance better than others. Overall, \textbf{\cref{tab:arfn-fr}} verifies the idea of using attention mechanisms to refine features that facilitate accurate detection of infrared small targets. We also test the performance of TBSG without feature concatenation and BOSE without element-wise feature addition, as shown in \textbf{\cref{tab:arfn-fr}} rows 1 and 7. The performance drops when features are not fused together to jointly represent structures or semantics, suggesting that in our case, a fused feature can better guide feature representations in our network.

\begin{table}[t]
    \centering
    \caption{Quantitative results when using different feature refinements in ARFN. Details can be found in \cref{sec:x-refine}.}
    % line \hyperlink{line-1006}{1006}-1021.}
    \begin{tabular}{l|c|c|c}
        \Xhline{1.2pt}
        \centering{\makecell[c]{\ \ \ \ \ \ Strategy}} & P$_{\uparrow}$ & R$_{\uparrow}$ & F-1$_{\uparrow}$ \\
        \hline
        TBSG w/o concatenation & 90.10 & 89.17 & 89.63 \\
        TBSG w/o channel refinement & 91.38 & 89.06 & 90.21 \\
        TBSG+CBAM \cite{x-woo2018cbam} & 91.19 & 90.74 & 90.96 \\
        TBSG+ECA Block \cite{x-wang2020eca} & 91.66 & 92.57 & 92.11 \\
        TBSG+LCDC \cite{x-fang2023danet} & 92.10 & 92.43 & 92.26 \\
        \rowcolor{gray!80} \textbf{TBSG+CR (Ours)} & \textbf{92.08} & \textbf{92.64} & \textbf{92.36} \\
        \hline
        BOSE w/o addition & 89.16 & 88.51 & 88.84 \\
        BOSE w/o spatial refinement & 88.57 & 90.04 & 89.30 \\
        BOSE+CBAM \cite{x-woo2018cbam} & 89.90 & 89.98 & 89.94 \\
        BOSE+CDSA \cite{x-fang2023dagnet} & 91.33 & 90.59 & 90.96 \\
        BOSE+LCDC \cite{x-fang2023danet} & 90.28 & 90.71 & 90.49 \\
        \rowcolor{gray!80} \textbf{BOSE+DSR (Ours)} & \textbf{92.08} & \textbf{92.64} & \textbf{92.36} \\
        \Xhline{1.2pt}
    \end{tabular}
    \vspace{-0.5em}
    \label{tab:arfn-fr}
\end{table}

\subsubsection{Different Approaches of Regional Feature Representation in LCM} \label{sec:x-lcm}

\begin{figure*}[!htbp]
    \centering
    \includegraphics[width=\linewidth,keepaspectratio]{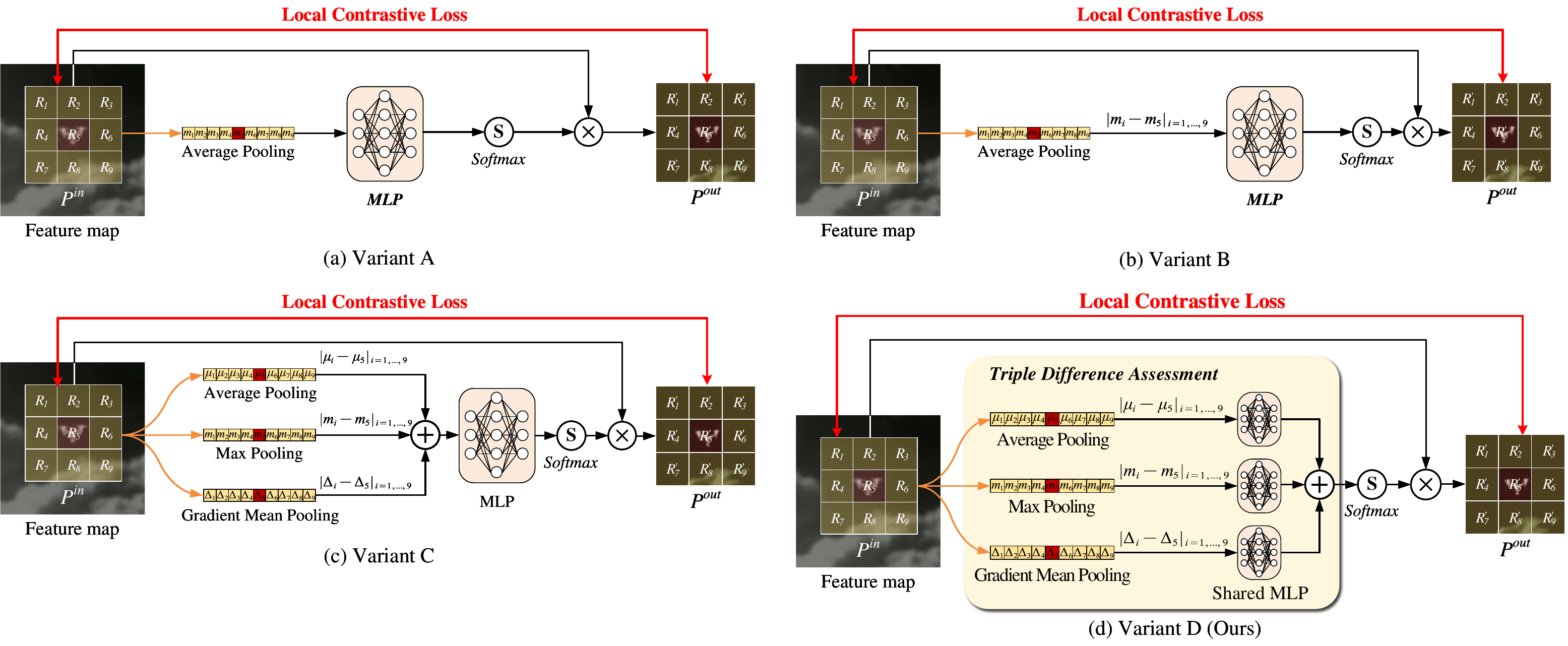}
    \vspace{-1.5em}
    \caption{Different variants of our LCM, including whether to use the feature difference calculation (FD), multi-branch regional feature representation (MB), or the use of shared MLP (SM). \cref{tab:lcm_variant} shows that our proposed variant D achieves the best performance. Details can be found in \cref{sec:x-lcm}.}
    % line \hyperlink{line-1034}{1034}-1043.}
    \vspace{-0.5em}
    \label{fig:lcm_structure}
\end{figure*}

\begin{table}[t]
    \centering
    % \footnotesize
    \caption{Effectiveness of various components in our LCM. Details can be found in \cref{sec:x-lcm}.}
    % line \hyperlink{line-1034}{1034}-1043.}
    \begin{tabular}{c|c|c|c|c|c|c}
        \Xhline{1.2pt}
        \multirow{2}*{Variant} & \multicolumn{3}{c|}{Components} & \multirow{2}*{P$_{\uparrow}$} & \multirow{2}*{R$_{\uparrow}$} & \multirow{2}*{F-1$_{\uparrow}$} \\
        \cline{2-4}
         & FD & MB & SM & & & \\
        \hline
        1 & - & - & - & 88.61 & 90.36 & 89.48 \\
        2 & $\checkmark$ & - & - & 90.69 & 90.45 & 90.57 \\
        3 & $\checkmark$ & $\checkmark$ & - & 91.29 & 90.81 & 91.05 \\
        \rowcolor{gray!80} \textbf{4 (Ours)} & $\checkmark$ & $\checkmark$ & $\checkmark$ & \textbf{92.08} & \textbf{92.64} & \textbf{92.36} \\
        \Xhline{1.2pt}
    \end{tabular}
    \vspace{-0.5em}
    \label{tab:lcm_variant}
\end{table}

\hypertarget{line-1034}{} In this section, we investigate how different approaches in extracting regional feature representations in LCM can affect the overall detection performance, e.g., for discriminating distractors. We examine three main factors of our proposed LCM: the feature difference calculation (FD), multi-branch regional feature representation (MB), and the use of shared MLP (SM), as shown in \textbf{\cref{fig:lcm_structure}}. From (a) to (d), we name each variant from A to D, each stands for a unique design of our LCM, and the detection performances of incorporating them into our CCDNet are reported in \textbf{\cref{tab:lcm_variant}}. We can see that each component in our proposed LCM can improve detection performance for CCDNet, while with every component employed in our LCM, the performance of our CCDNet achieves the best, validating the effectiveness of the design of our LCM. It is noteworthy that our proposed LCM can improve metric P more than matric R, with an improvement of 3.41\% in P and 1.12\% in R. This is because our LCM is designed to discriminate distractors in the complex backgrounds, which serves to reduce the false alarm rate and thus improve metric P.

\begin{figure*}[!htbp]
    \centering
    \includegraphics[width=0.85\linewidth,keepaspectratio]{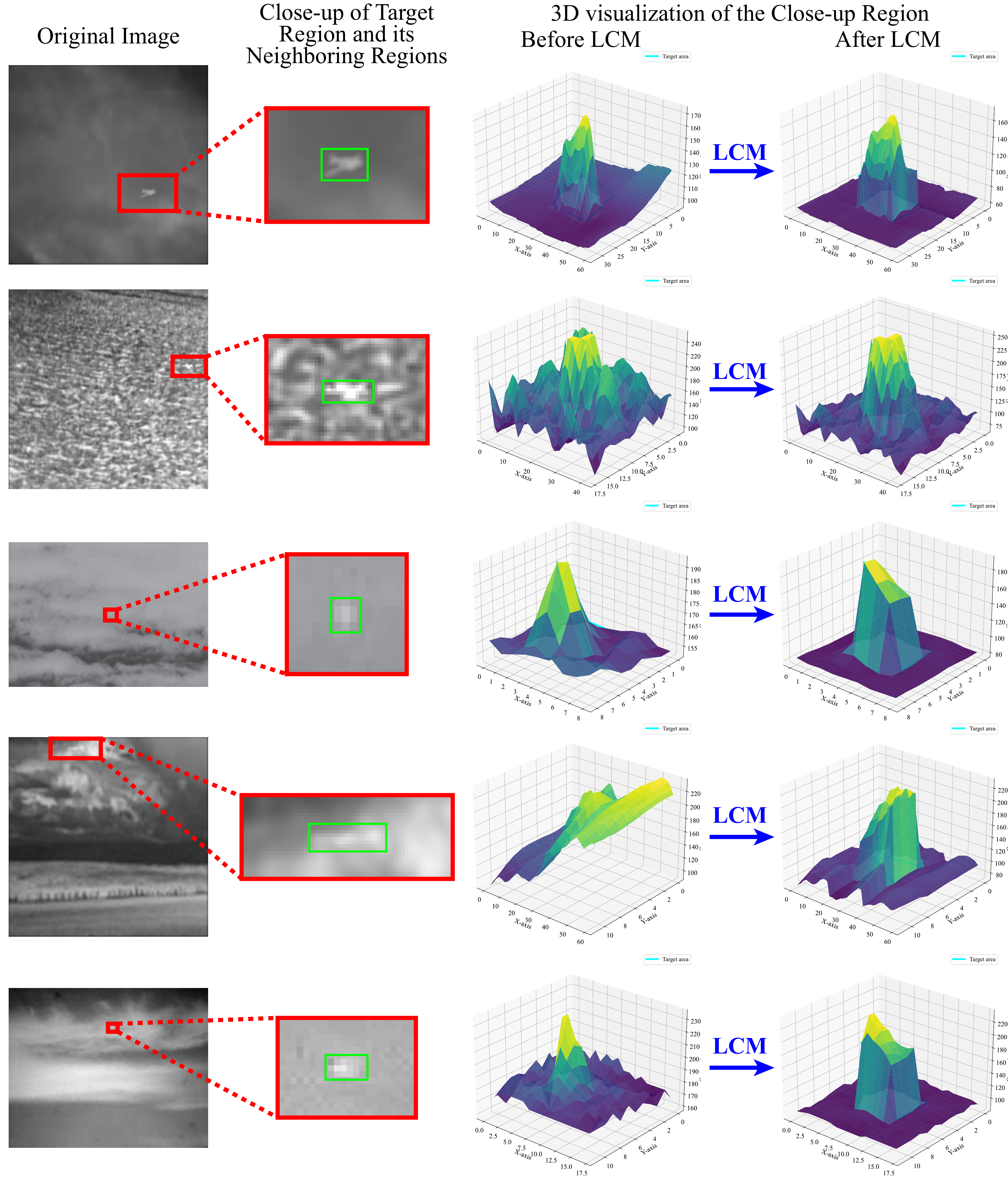}
    \caption{Illustration and Visualization of the impact of our proposed LCM in training stage. The results show that our proposed LCM can effectively highlight the target region and suppress its surrounding regions, which can facilitate an accurate detection performance. Details can be found in \cref{sec:x-lcm}.}
    % line \hyperlink{line-1044}{1044}-1071.}
    \vspace{-1em}
    \label{fig:lcm_visual}
\end{figure*}

\hypertarget{line-1044}{} To illustrate the impact of our LCM during network training, we display the 3D visualization plots of feature maps of the targets and their neighboring regions to demonstrate how our LCM can improve the saliency of the targets while suppressing the surroundings. As shown in \textbf{\cref{fig:lcm_visual}} column 3, target features before LCM are usually surrounded by relatively ``uneven'' background features, especially for row 2 and row 5. In this situation, targets might potentially be camouflaged in their surrounding backgrounds, and thus become difficult to identify. After we adopt LCM and enforce contrastive operations on them, a clear contrast between the target and the background features can be seen in column 4, as the background features are evidently suppressed compared to the highlighted foreground. As targets possess a stark contrast to their neighboring regions, it will be easier to identify since they have higher saliency. In particular, for rows 2 and 5, the backgrounds are aggressively suppressed and the targets can be seen clearly. Therefore, these visualizations have proven that our proposed LCM is able to instill contrastive knowledge of target features in our CCDNet via local target-background contrastive learning.

It is worth noticing that, although our proposed LCM achieves great performance in our CCDNet, it is only used during network training. This means our LCM only serves as a penalty to network training instead of directly enhancing target regions as in \textbf{\cref{fig:lcm_visual}} during inference time. How to construct an adaptive feature enhancing module with the same effectiveness as our LCM for inference can be a research direction in the future.

\begin{figure}[t]
    \centering
    \includegraphics[width=\linewidth,keepaspectratio]{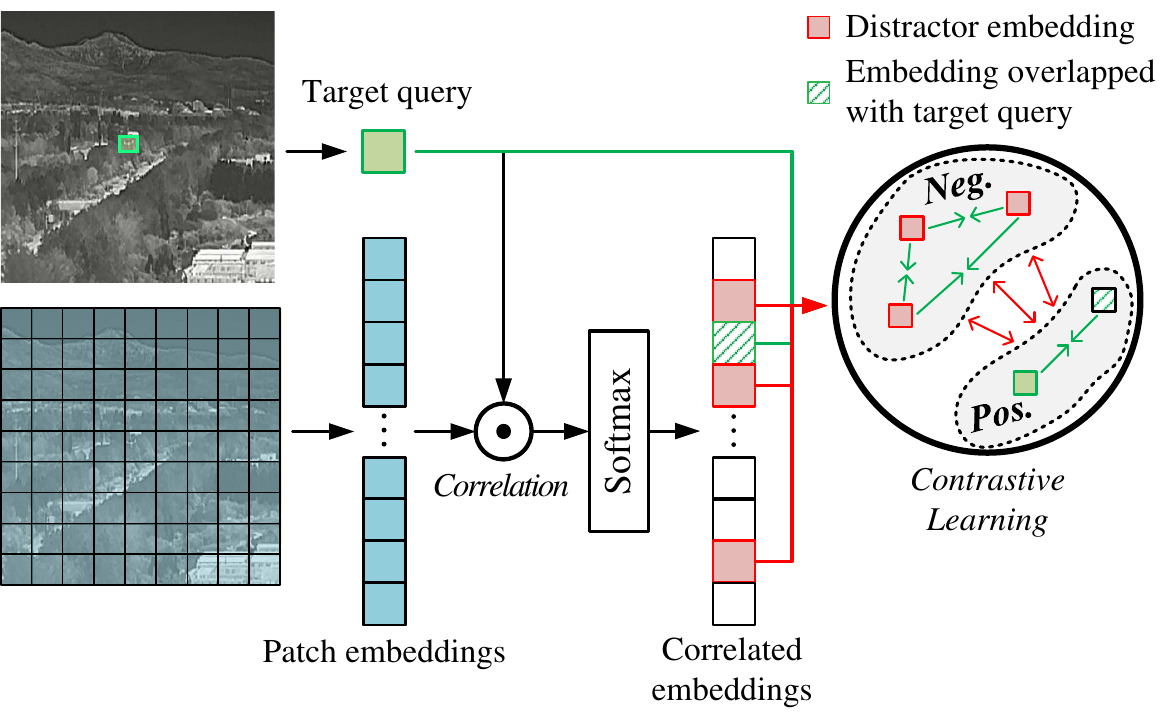}
    \vspace{-1em}
    \caption{The structure of cross-attention GCM (CA-GCM), another adaptive global contrastive module we design for ablation study. Details can be found in \cref{sec:x-gcm}.}
    % line \hyperlink{line-1079}{1079}-1086.}
    \vspace{-0.5em}
    \label{fig:another-gcm}
\end{figure}

\subsubsection{Discussion on the Design of GCM} \label{sec:x-gcm}

\hypertarget{line-1073}{} \ \ \ \ \textbf{Different global contrastive learning structure}. In our proposed GCM, the positive and negative samples are obtained through an adaptive sample mining mechanism (in \textbf{\cref{sec:3.4.2}}). \hypertarget{line-1079}{} We study this adaptive mechanism and compare it to other measures (e.g., ADCR \cite{x-fang2023danet}). The key challenge is to define the negative samples, which are object features similar to the real targets. We intuitively design another approach to achieve this, using cross-attention similarity to find distractors as negative samples, as shown in \textbf{\cref{fig:another-gcm}}. This cross-attention-based GCM (CA-GCM) patchifies the input feature map into different patches, then converts them into feature embeddings, and performs cross-attention with the target feature query to find the top-k (k=3) related embeddings as negative samples. We employ these contrastive learning modules in our CCDNet and conduct ablation studies, and the performances are reported in \textbf{\cref{tab:gcm-quantitative}}.

We can see from \textbf{\cref{tab:gcm-quantitative}} that our proposed GCM achieves the best performance for CCDNet. This is because our GCM takes the detection results into account, regarding those with lower confidence scores as negative samples that our proposed network struggles to identify. Comparatively, CD-GCM achieves the second-best performance. We hypothesize that the top-k embedding selection might not be suitable for negative sample mining, since it only selects highly correlated embeddings as negative samples, which can be regarded as hard negative samples. However, only having hard samples might be a detriment to effective contrastive learning since it might bring too much difficulty in discriminating with positive samples, thus yielding suboptimal detection performance. And for ADCR, its random negative sample mining strategy will lead to confusion in contrastive learning, and so have limited performance improvement.

\begin{table}[t]
    \centering
    \caption{Performance of our CCDNet employed with different global contrastive modules. Details can be found in \cref{sec:x-gcm}.}
    % line \hyperlink{line-1073}{1073}-1089.}
    % \vspace{-0.5em}
    \begin{tabular}{l|c|c|c}
        \Xhline{1.2pt}
        \makecell[c]{\ \ \ \ \ \ Model} & P$_{\uparrow}$ & R$_{\uparrow}$ & F-1$_{\uparrow}$ \\
        \hline
        CCDNet w/o GCM & 88.89 & 90.98 &  89.92\\
        \rowcolor{gray!80} \textbf{CCDNet+GCM (Ours)} & \textbf{92.08} & \textbf{92.64} & \textbf{92.36} \\
        CCDNet+CA-GCM & \ul{89.74} & \ul{91.66} & \ul{91.20} \\
        CCDNet+ADCR \cite{x-fang2023danet} & 89.09 & 91.75 & 90.40 \\
        \Xhline{1.2pt}
    \end{tabular}
    % \vspace{-0.5em}
    \label{tab:gcm-quantitative}
\end{table}

\begin{table}[t]
    \centering
    \caption{Impact of different thresholds on our GCM. $t_1$ and $t_2$ are two confidence thresholds introduced in \cref{eq:gcm-threshold}, respectively. Details can be found in \cref{sec:x-gcm}.}
    % line \hyperlink{line-1107}{1107}-1129.}
    % \vspace{-0.5em}
    \begin{tabular}{c|c|c|c|c}
        \Xhline{1.2pt}
        $t_1$ & $t_2$ & P$_{\uparrow}$ & R$_{\uparrow}$ & F-1$_{\uparrow}$ \\
        \hline
        0.9 & 0.2 & 91.88 & \ul{92.54} & \ul{92.21} \\
        \rowcolor{gray!80} \textbf{0.8} & \textbf{0.2} & \textbf{92.08} & \textbf{92.64} & \textbf{92.36} \\
        0.7 & 0.2 & 89.63 & 92.09 & 90.84 \\
        0.6 & 0.2 & 87.92 & 90.65 & 89.26 \\
        0.5 & 0.2 & 86.17 & 90.84 & 88.45 \\
        \hline
        0.8 & 0.3 & \ul{91.89} & 91.65 & 91.77 \\
        0.8 & 0.4 & 90.13 & 89.19 & 89.66 \\
        0.8 & 0.5 & 87.78 & 89.28 & 88.52 \\
        \Xhline{1.2pt}
    \end{tabular}
    % \vspace{-1em}
    \label{tab:gcm-threshold}
\end{table}

\hypertarget{line-1107}{} \textbf{Impact of different sample definition threshold pairs}. As given in \cref{eq:gcm-threshold}, when a detection result has a confidence score larger than $t_1$, it is assigned as a positive sample; if it is between $t_1$ and $t_2$, it will be a negative sample; otherwise, it will not be considered for use in global contrastive learning. We ablate different pairs of thresholds ($t_1$ and $t_2$ in \cref{eq:gcm-threshold}) to see how they affect the overall detection performance, as reported in \textbf{\cref{tab:gcm-threshold}}. 

\begin{figure*}[t]
    \centering
    \includegraphics[width=\linewidth,keepaspectratio]{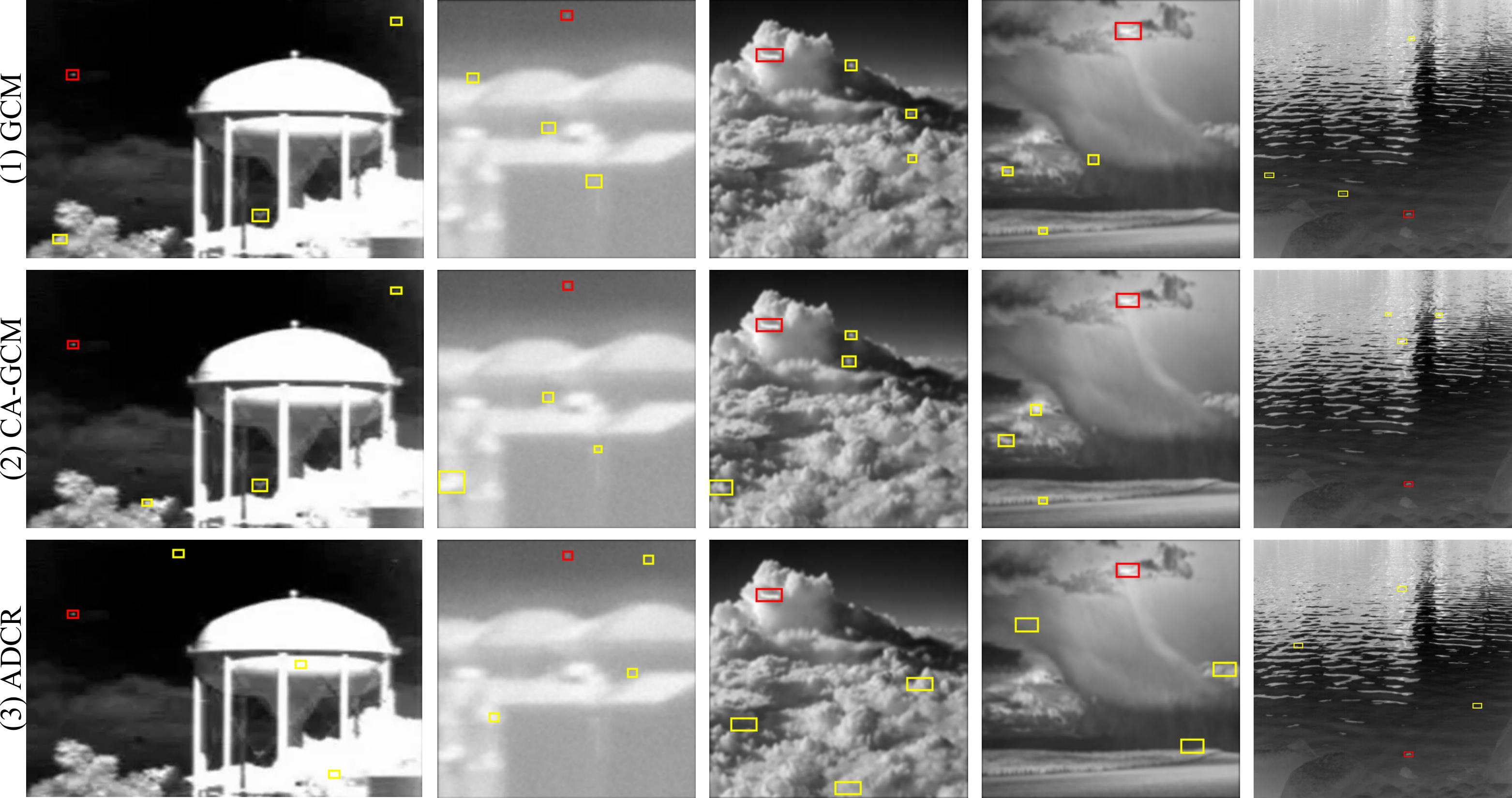}
    \caption{Illutrations of positive samples (\textcolor{red}{red boxes}) and negative samples (\textcolor{yellow}{yellow boxes}) selected in different global contrastive learning modules during network training. Details can be found in \cref{sec:x-gcm}.}
    % line \hyperlink{line-1129}{1129}-1145.}
    \vspace{-1em}
    \label{fig:gcm-pics}
\end{figure*}

As can be seen, when the positive sample threshold drops from 0.8 to 0.5, detection performance decreases from 92.08 to 86.17 in P and 92.36 to 88.45 in F-1. This suggests that a lower positive sample threshold might introduce too many positive samples, many of which might not be correct detection results. However, as we increase the positive sample threshold to 0.9, the detection performance drops again, which we believe is due to fewer positive samples for contrastive learning. On the other hand, the detection performance drops as we increase the negative sample threshold. Fewer negative samples are selected when the threshold is raised, leading to suboptimal contrastive learning performance. Therefore, we select 0.8 and 0.2 as our positive and negative sample thresholds, respectively.

It is noteworthy that, the purpose of our GCM is to differentiate real targets from hard ambiguous distractors with similar features. Generally, detections of confidence lower than $t_2$ are considered easy negatives as they may represent pure backgrounds, and those greater than $t_1$ are easy positives since they contain evident target features. Distractors are mostly hard samples with confidence scores between $t_1$ and $t_2$. Since easy positives (conf$>t_1$) and easy negatives (conf$\le t_2$) can be well identified, our proposed GCM is only applied on hard samples with confidence between $t_2=0.2$ and $t_1=0.8$ to learn discriminative features for real targets and distractors distinction.

\hypertarget{line-1129}{} \textbf{Illustration of sample selection}. We display some sample selection examples for different global contrastive learning modules in \textbf{\cref{fig:gcm-pics}}, where red boxes are the target regions as positive samples, and yellow regions are the distractor regions as negative samples. As shown in \textbf{\cref{fig:gcm-pics}}, the GCM and CA-GCM extract negative samples more effectively than ADCR, which selects negative samples randomly on the feature maps according to the target bounding boxes. In contrast to CA-GCM, our GCM extracts negative sample regions more related to the targets. For example, in the last column of rows (1) and (2), CA-GCM extracts more regions of the bright waves, deviating from the target characteristics. Our GCM focuses more on distractors that have intrinsically similar features to the real targets via adaptively utilizing the network's detection results, thus selecting more interpretable negative samples and facilitating more to the detection performance.

\subsection{Qualitative Results for the Rest Comparisons} \label{sec:x-restqua}

\begin{figure*}[!htbp]
    \centering
    \includegraphics[width=0.85\linewidth,keepaspectratio]{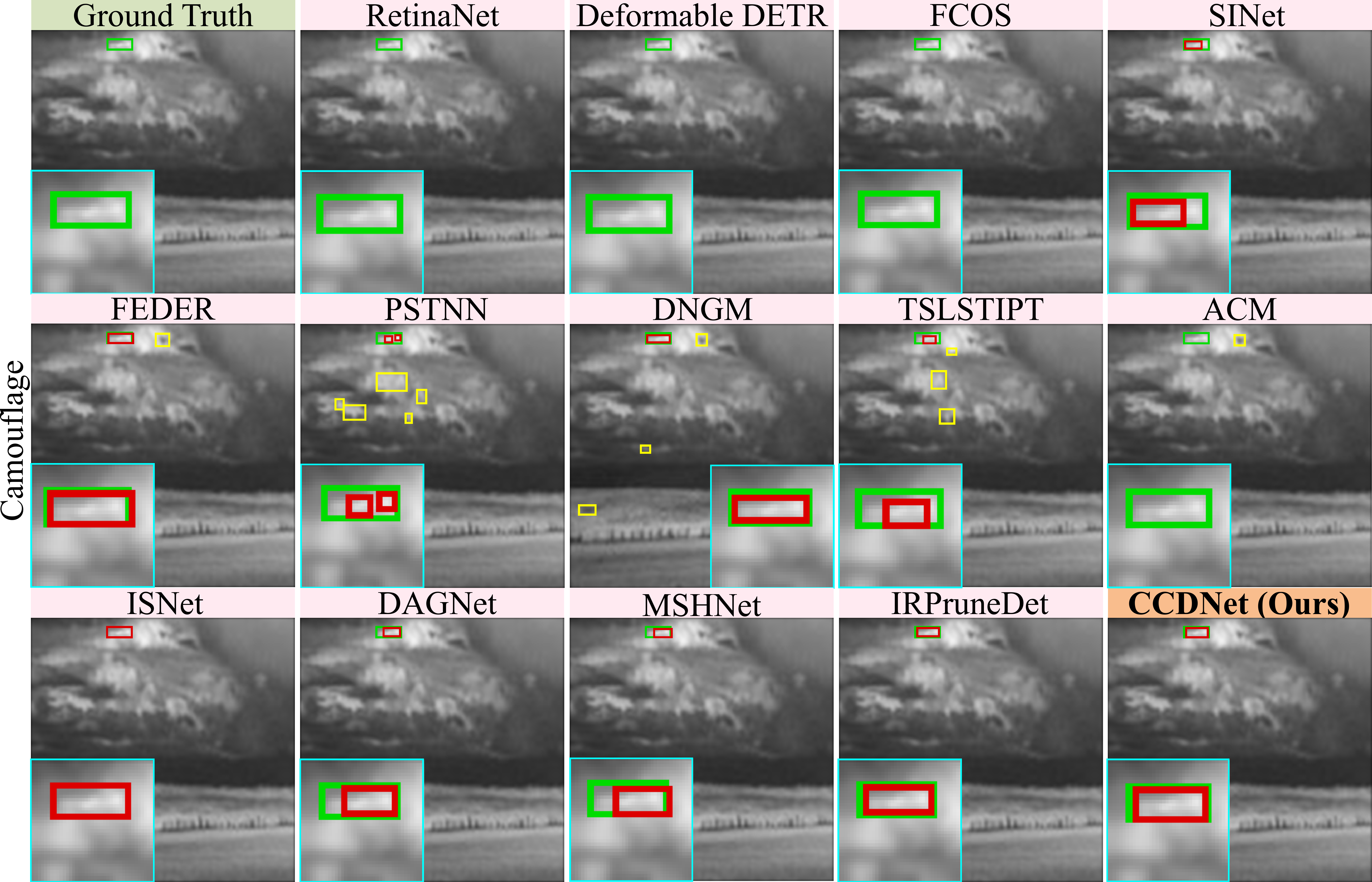}
    \caption{Qualitative results of all the comparison methods regarding a target camouflage scenario. The \textcolor{red}{red}, \textcolor{green}{green}, and \textcolor{yellow}{yellow} boxes are ground truths, detection results, and false alarms. An image without any red box indicates a missed detection. We put close-ups for each detection result for better visual comparison.}
    \label{fig:supple-results-1}
\end{figure*}

\begin{figure*}[!htbp]
    \centering
    \includegraphics[width=0.85\linewidth,keepaspectratio]{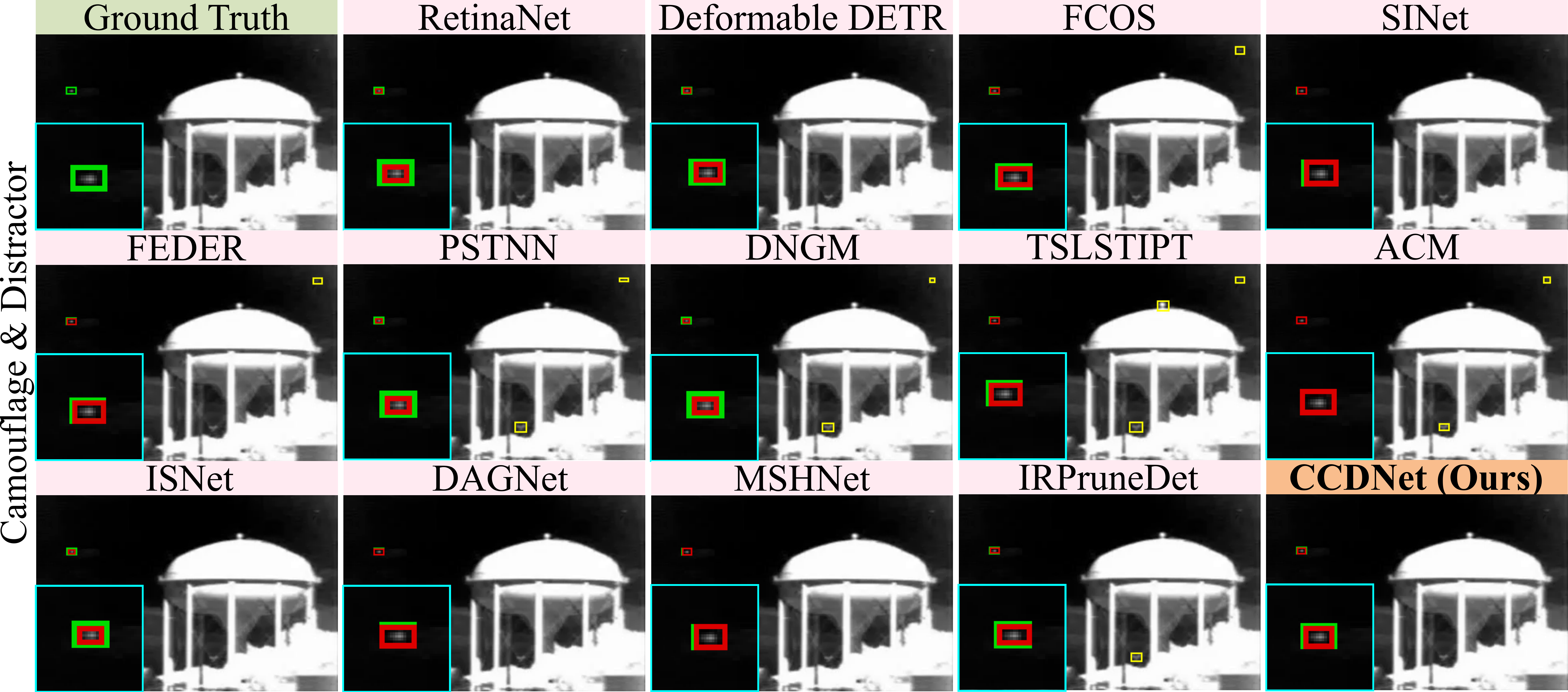}
    \caption{Qualitative results of all the comparison methods regarding a target camouflage scenario. The \textcolor{red}{red}, \textcolor{green}{green}, and \textcolor{yellow}{yellow} boxes are ground truths, detection results, and false alarms. An image without any red box indicates a missed detection. We put close-ups for each detection result for better visual comparison.}
    \vspace{-1em}
    \label{fig:supple-results-2}
\end{figure*}

\begin{figure*}[!htbp]
    \centering
    \includegraphics[width=0.85\linewidth,keepaspectratio]{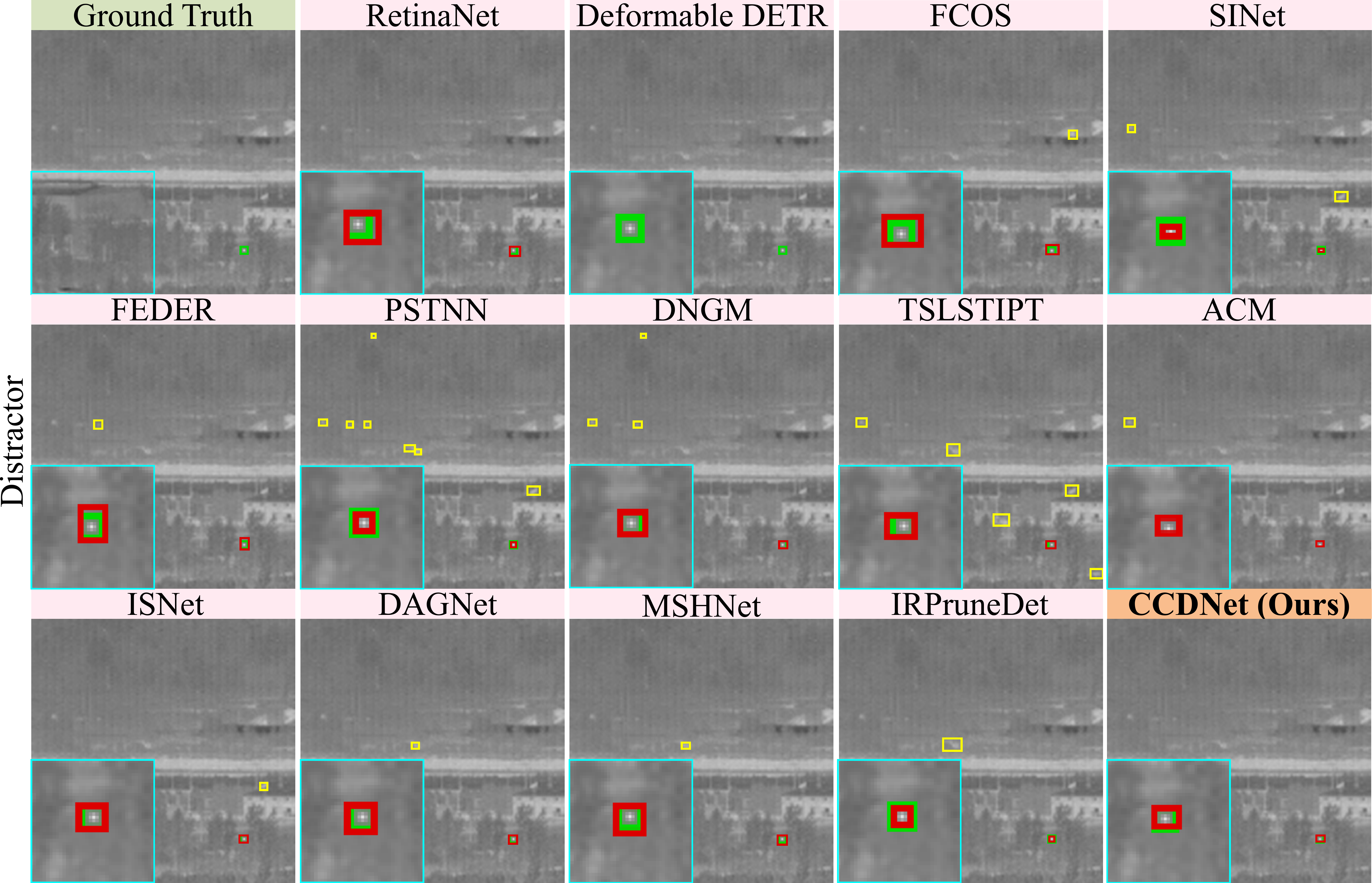}
    \caption{Qualitative results of all the comparison methods regarding a target camouflage scenario. The \textcolor{red}{red}, \textcolor{green}{green}, and \textcolor{yellow}{yellow} boxes are ground truths, detection results, and false alarms. An image without any red box indicates a missed detection. We put close-ups for each detection result for better visual comparison.}
    \label{fig:supple-results-3}
\end{figure*}

\begin{figure*}[!htbp]
    \centering
    \includegraphics[width=0.85\linewidth,keepaspectratio]{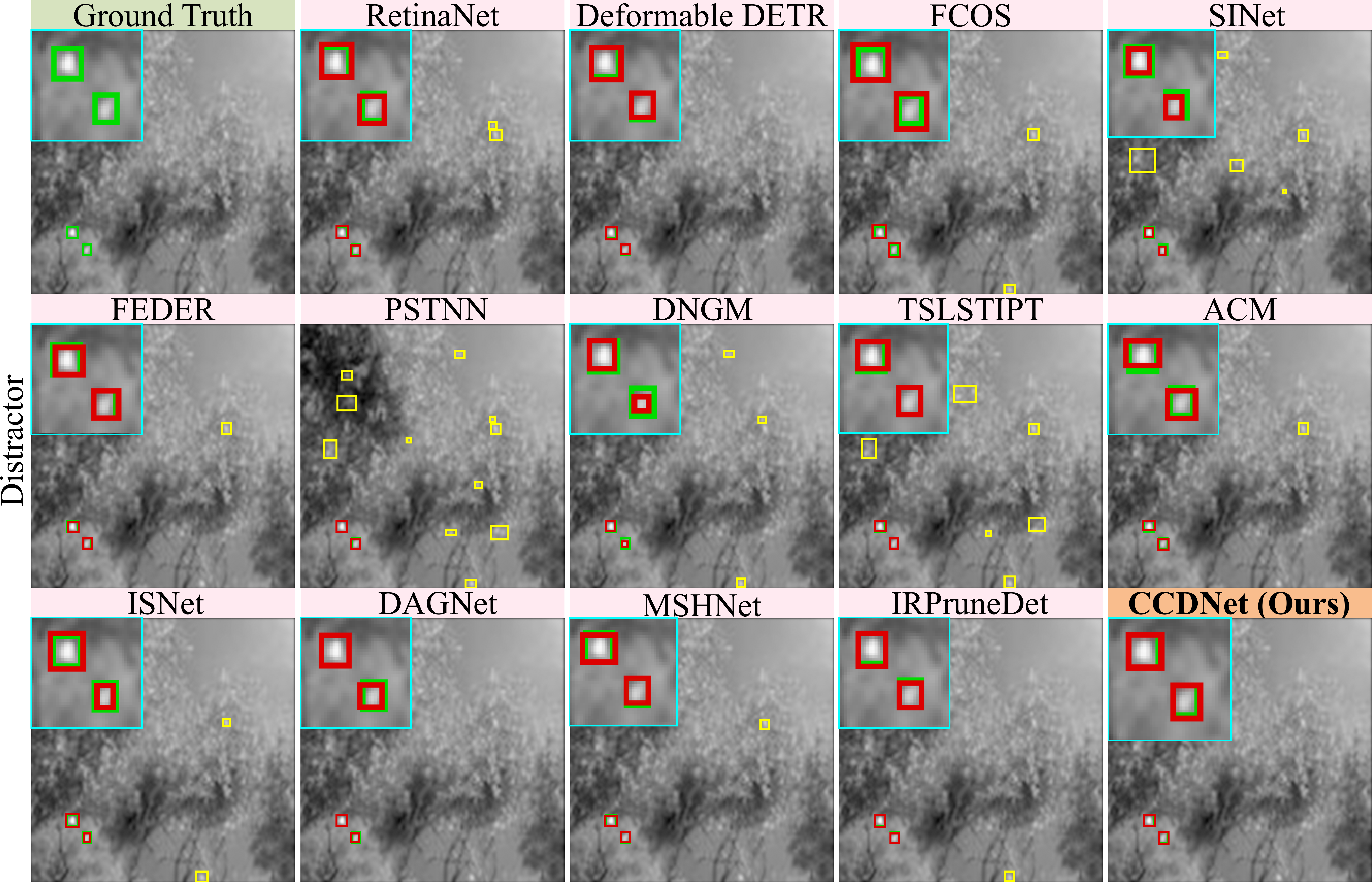}
    \caption{Qualitative results of all the comparison methods regarding a target camouflage scenario. The \textcolor{red}{red}, \textcolor{green}{green}, and \textcolor{yellow}{yellow} boxes are ground truths, detection results, and false alarms. An image without any red box indicates a missed detection. We put close-ups for each detection result for better visual comparison.}
    \label{fig:supple-results-4}
\end{figure*}

In our main paper, we only display qualitative results from five different comparison methods due to space limits. In this section, we present the qualitative results of all the comparison methods in \textbf{\cref{fig:supple-results-1,fig:supple-results-2,fig:supple-results-3,fig:supple-results-4}}. As we can see, compared to all comparison methods, our proposed CCDNet achieves the most consistent detection results with no false alarms and missed detections. At the same time, the detection bounding boxes of our CCDNet overlap comparatively better with the ground truths, verifying the effectiveness of our CCDNet.

{
    \small
    \bibliographystyle{ieeenat_fullname}
    \bibliography{main}
}

\end{document}